\documentclass[sigconf]{acmart}
\usepackage{bbm}
\usepackage{amsmath}
\usepackage{amsthm}
\usepackage{amsfonts}

\AtBeginDocument{%
  \providecommand\BibTeX{{%
    \normalfont B\kern-0.5em{\scshape i\kern-0.25em b}\kern-0.8em\TeX}}}

\begin{document}

\title{Mix-and-Match: Scalable Dialog Response Retrieval using Gaussian Mixture Embeddings}

%
\author{Gaurav Pandey}
\email{gpandey1@in.ibm.com}
\affiliation{%
  \institution{IBM Research}
  \country{India}
}
\author{Danish Contractor}
\email{dcontrac@in.ibm.com}
\affiliation{%
  \institution{IBM Research}
  \country{India}
}
\author{Sachindra Joshi}
\email{jsachind@in.ibm.com}
\affiliation{%
  \institution{IBM Research}
  \country{India}
}

\renewcommand{\shortauthors}{Trovato and Tobin, et al.}
\newcommand{\cc}{\mathbf{c}}
\newcommand{\rr}{\mathbf{r}}
\newcommand{\T}{\mathrm{T}}
\newcommand{\HH}{\mathcal{H}}
\newcommand{\KL}{\mathrm{KL}}
\newcommand{\dz}{\mathrm{d}z}

\begin{abstract}
Embedding-based approaches for dialog response retrieval embed the context-response pairs as points in the embedding space. These approaches are scalable, but fail to account for the complex, man-to-many relationships that exist between context-response pairs. On the other end of the spectrum, there are approaches that feed the context-response pairs jointly through multiple layers of neural networks. These approaches can model the complex relationships between context-response pairs, but fail to scale when the set of responses is moderately large (>100). In this paper, we combine the best of both worlds by proposing a scalable model that can learn complex relationships between context-response pairs. Specifically, the model maps the contexts as well as responses to probability distributions over the embedding space. We train the models by optimizing the Kullback-Leibler divergence between the distributions induced by context-response pairs in the training data. We show that the resultant model achieves better performance as compared to other embedding-based approaches on publicly available conversation data.\end{abstract}


\maketitle

\section{Introduction}
\label{sec:introduction}
\begin{figure}
    \centering
    \includegraphics[scale=0.55]{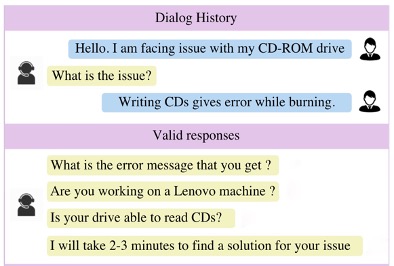}
    \caption{An example of a context with multiple valid responses. Note that each response contains different information and hence must have embeddings that are far way from each other. However, embedding-based approaches for retrieval attempt to bring all such responses close to the context and hence, close to each other. }
    \label{fig:one_to_many_example}
\end{figure}

Since the advent of deep learning, several neural network-based approaches have been proposed for predicting responses given a dialog context (the set of utterances so far). These models can broadly be classified into generative and retrieval-based. Generative response predictors feed the dialog context to an encoder (flat~\cite{Vinyals2015ANC,Sordoni2015ANN,zhang2020dialogpt} or hierarchical~\cite{Serban2016BuildingED}) and the resultant embeddings are fed to a decoder to generate the response token-by-token. When these models were deployed on real-world conversations, it was found that the generated responses were often uninformative and lacked diversity~\cite{diversitypaper}. To incorporate diversity among the responses, 
variants of this standard architecture that use latent variables have also been explored~\cite{serban2017hierarchical,cao2017latent, park2018hierarchical,gu2018dialogwae}.

In contrast to generative models, retrieval-based response predictors~\cite{Ji2014AnIR,Yan2016LearningTR,Wu2017ASM,Bartl2017ARD} retrieve the response from a predefined set of responses given the dialog context. Such methods find application in a variety of real-world dialog modeling and collaborative human-agent tasks. For instance, dialog modeling frameworks typically utilize the notion of ``intents'' and ``dialog flows'' which aim to model the ``goal'' of a user-utterance \cite{dialogframeworks}. To make task of building and identifying such intents easier, some tools mine conversation logs to identify responses that are often associated with dialog contexts (intents) \cite{AAAI21} and then surface these responses for review by humans. These reviewed responses are then modeled into the dialog flow for different intents.  Another instance, of human-agent collaboration powered by system returned responses is in `Agent Assist' environments where a system makes recommendations to a customer-support or contact-center agent in real-time\cite{AgentAssist}. 
Responses from retrieval based systems score higher on fluency and informativeness and have also been used to power real-world chat bots \cite{IJCAI21}.

The success of a good response retrieval system lies in learning a good similarity function between the context and the response.  In addition, it also needs to be scalable so that it can retrieve responses from the universe of tens of thousand of responses efficiently. These two requirements present a tradeoff between the richness of scoring and scalability, as discussed below. 

\noindent {\bf Trade-off between Scoring and Scalability: } Typically, in neural dialog retrieval models, the contexts and the responses in the conversation logs are embedded as points in the embedding space~\cite{lowe2015ubuntu}.  Approaches such as contrastive learning~\cite{bromley1993signature} are then used to ensure that the context is closer to the ground-truth response than the other responses. 
Figure~\ref{fig:one_to_many_example} shows a dialog context followed by multiple responses. Despite the apparent diversity among responses, all the responses are valid for the dialog context.  A typical embedding-based approach for retrieval would bring the embedding of the dialog context close to the embedding of all the valid responses~\cite{DPR,ConvDR,ANNICLR,RetrievalTACL}. 
However this has the undesirable effect of making the valid, but diverse, responses gravitate towards each other in the embedding space. Similarly, a generic response is a valid response for several dialog contexts. Again, an embedding-based approach would bring all the context embeddings close to the embedding of the generic response, even though the contexts are unrelated to each other. 

Thus, typical embedding-based approaches for retrieval fail to capture the complex, many-to-many relationships that exist in conversations. More complex matching networks such as Sequential Matching Networks~\cite{Wu2017ASM} and BERT~\cite{CoBERT} based cross-encoders jointly feed the context-response pairs through multiple layers of neural networks for generating the similarity score. While these approaches have proven to be effective for response retrieval, they are very expensive in terms of inference time. Specifically, if $N_c$ is the total number of dialog contexts and $N_r$ is the total number of responses available for retrieval during inference, these methods have a time complexity of $O(N_cN_r)$. Hence, they can't be used in a real-world setting for retrieving from thousands of responses.  

\noindent {\bf Contributions: } An effective response retrieval system must be able to capture the complex relationship that exists between a context and a response, while simultaneously being fast enough to be deployed in the real world. In this paper we present a scalable and efficient dialog-retrieval system that maps the contexts as well as the responses to probability distributions over the embedding space (instead of points in the embedding space). To capture the complex many-to-many relationships between the context and response, we use multimodal distributions such as Gaussian mixtures to model each context and response. The resultant model is referred to as `Mix-and-Match'. Intuitively, if a response is a valid response for a given dialog context, we want the corresponding probability distribution to be "close" to the context distribution.  We formalize this notion of closeness among distributions by using Kullback-Leibler (KL) divergence. Specifically, we minimize the Kullback-Leibler divergence between the context distribution and the distribution of the ground-truth response while maximizing the divergence from the distributions of other negatively-samples responses. We derive approximate but closed-form expressions for the KL divergence when the underlying distributions are Gaussian mixtures. This approximation significantly alleviates the computation cost of KL-divergence, thereby making it suitable for use in real-world settings. In addition, we state how our model reduces to some existing multi-embedding representations~\cite{khattab2020colbert} under certain assumptions about the nature of Gaussian Mixtures.  We  demonstrate our work on two publicly available dialog datasets  -- Ubuntu Dialog Corpus (v2)\cite{lowe2015ubuntu} and the Twitter Customer Support dataset\footnote{ https://www.kaggle.com/thoughtvector/customer-support-on-twitter} as well as on an internal real-world technical support dataset. Using automated as well as human studies, we  demonstrate that Mix-and-Match outperforms recent embedding-based retrieval methods. 

\section{Related Work}
Our work is broadly related with two current areas of research - Resonse retrieval (Section \ref{sec:related-response} and Probabilitic Embeddings (Section \ref{sec:related-probemb}).
\subsection{Response Retrieval Systems} \label{sec:related-response}
Retrieval based systems for dialog models have been applied in a variety of settings. Existing work has studied the problem of grounding responses in external knowledge such as documents~\cite{OrQUAC,RAG,VRAG}, structured knowledge \cite{Multilevel,MWozBiswesh}, with varying degrees of knowledge-level supervision~\cite{Multilevel,DineshTACL}. In such cases, a knowledge instance is first fetched and then a response is generated. In contrast to knowledge grounded responses, in response retrieval settings, the dialog context is used to directly fetch responses from a universe of responses without relying on external knowledge. Depending on how the context and responses are encoded for retrieval, approaches can be classified into methods that use: (i) Independent encodings (ii) Joint Encodings.

\noindent {\bf Independent Encodings: } One of the earliest methods used for dialog retrieval uses TF-IDF \cite{BM25} scores to represent context and responses. A common architecture employed by neural methods for dialog retrieval is a dual encoder. Here, the context and responses are encoded using a shared architecture but in different parameter spaces. Early versions of such methods employed LSTMs \cite{lowe2015ubuntu} but more recently, pre-trained models have been used \cite{DPR,lu2020twinbert,SBERT,liu-etal-2021-quadrupletbert}. Models such as DPR \cite{DPR}, S-BERT \cite{SBERT}, MEBERT \cite{MEBERT} encode contexts and responses using dual encoders based on the BERT \cite{BERT} pre-trained model, and learn a scoring function using negative samples. Work that focuses on improving re-ranking by selecting better negative samples has also been done \cite{ANNICLR,liu-etal-2021-quadrupletbert}. Models such as Poly-Encoder \cite{PolyEncoder}, MEBERT \cite{MEBERT}, ColBERT\cite{khattab2020colbert} use multiple representations for dialog contexts instead of using a single representation. While PolyEncoder generates multiple encodings for the dialog context using a special attention layer, MEBERT\cite{MEBERT} directly creates multiple embedding representations using specialized layers. However, instead of using multiple encoders, CoLBERT uses a BERT based dual-encoder architecture to encode contexts and responses,\footnote{The work was originally presented for retrieval in QA tasks} but it does not use a single embedding for scoring. Instead, it uses a scoring function that directly operates on the contextual token representations of the dialog context and responses. In particular, it uses the maximum similarity between any contextual representation pair to compute the overall similarity. The advantage of these approaches is that it makes the similarity function more expressive while retaining the the scalability offered by traditional dual-encoder architectures. 
\begin{figure*}[ht]
    \centering
    \includegraphics[scale=0.38]{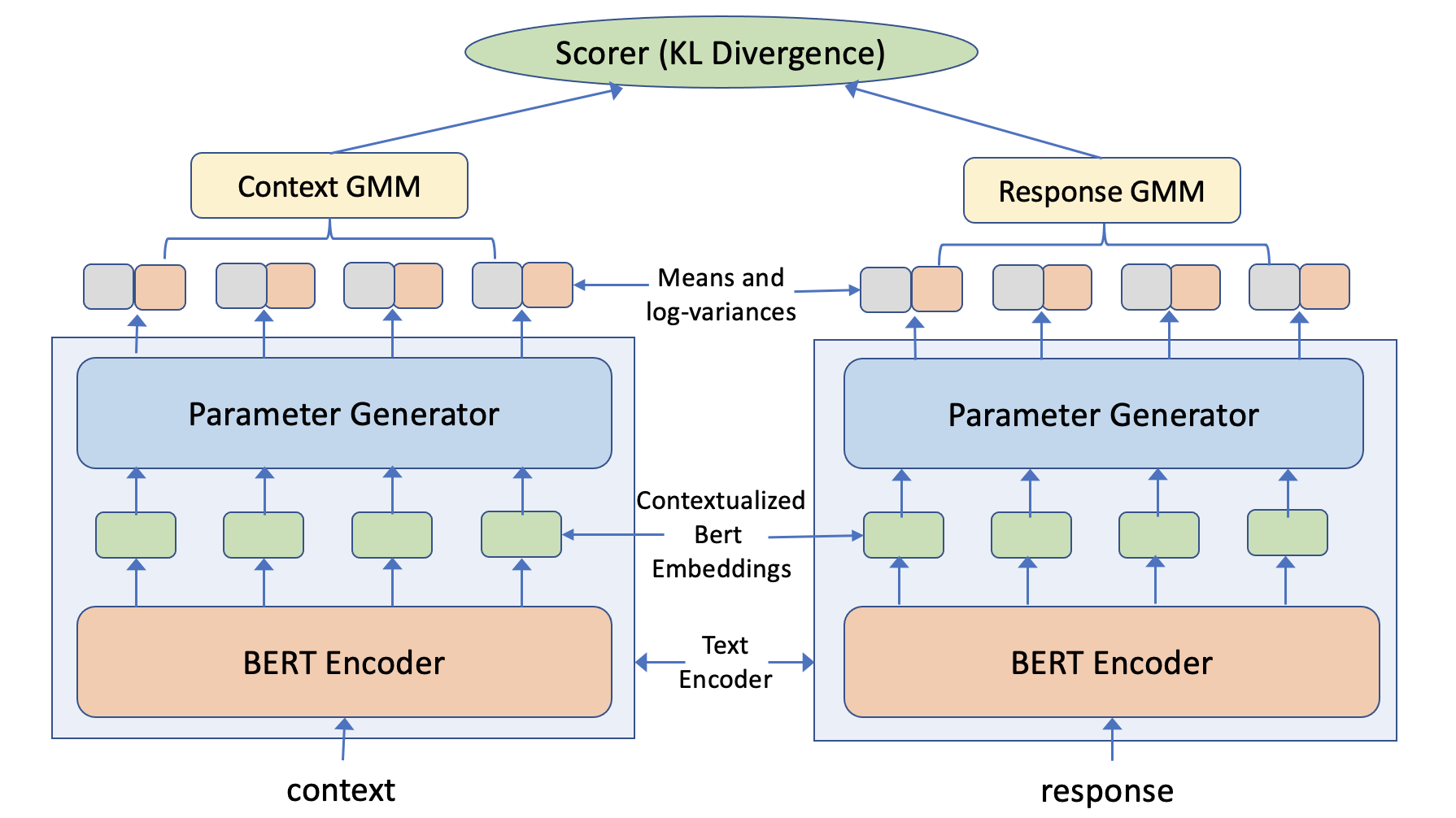}
    \caption{An overview of our model -  Mix-and-Match. }
    \label{fig:model}
\end{figure*}

\noindent {\bf Joint Encoding: }
In contrast to methods that independently encode context and response pairs, methods such as Sequential Matching Networks~\cite{Wu2017SequentialMN}, Cross encoders using BERT \cite{BERTIR,CoBERT} jointly encode context and dialog responses. However, such models are slow during inference because all candidate responses need to be jointly encoded with the dialog context for scoring at runtime. This is in contrast to dual-encoder architectures where response embeddings can be computed offline and cached for efficient retrieval. Models such as ConvRT \cite{distillconvIR}, TwinBERT \cite{lu2020twinbert} use distillation to train a dual encoder from a cross encoder models to help a train better dual-encoder model. 
\subsection{Probabilistic Embeddings} \label{sec:related-probemb}
Probabilistic embeddings have been applied in tasks for building better word representations \cite{GaussianEMB2,GaussianEmb}, entity comparison \cite{entcmp}, facial recognition \cite{facialreco}, pose estimation \cite{poseprob}, generating multimodal embeddings \cite{athiwaratkun-wilson-2017-multimodal,Chun2021ProbabilisticEF}, etc. The motivation in some of these tasks is similar to ours -- for instance, \citet{GaussianEMB2} use Gaussian embeddings to represent words to better capture meaning and ambiguity. However, to the best of our knowledge, the problem of applying probabilistic embeddings in dialog modeling tasks hasn't been explored. In this work, we represent dialog contexts as Mixture of Gaussians  present approximate closed form expressions for efficiently computing KL-divergence based distance measures, thereby making it suitable for use in real-world settings.

\section{Mix-and-Match}
To capture the complex many-to-many relationships between the dialog context and a response, we use  Gaussian mixtures to model each context and response. We consider a dialog to be a sequence of utterances $(u_1, \ldots, u_n)$. At any time-step $t$, the set of utterances prior to that time-step is referred to as the context. 
The utterance that immediately follows the context\footnote{We use the words context and dialog context interchangeably throughout the paper.} is referred to as the response. 
Instead of modeling the context and response as point embeddings, we use probability distributions induced by the context and the response on the embedding space,  denoted as $p_c(z)$ and $p_r(z)$\footnote{Formally, these are densities induced by the corresponding distributions} respectively, where $z$ is any point in the embedding space $\mathbb{R}^d$.


\subsection{Overview}
An overview of the model is shown in Figure~\ref{fig:model}. The context and response are first encoded using a pre-trained BERT model. The model consists of a Gaussian Mixture Parameter Generator, $\pi(X,K)$, which takes as input an encoded text sequence $X$ along with the number of Gaussian Mixtures, $K$ and then returns the means $\mu_k$ and variance $\sigma_k^2$ for the every Gaussian mixture component $k \in \{1, \dots K\}$, as its output. The encoded representations of the context and response from BERT are used to generate Gaussian Mixture distributions over the embedding space $\mathbb{R}^d$ using the parameter generator $\pi$.   
We then compute the $\KL$ divergence between the context and response distributions and use contrastive loss to bring the context closer to the ground-truth response as compared to other, negatively-sampled responses. 


\subsection{Text Encoder}
The text encoder maps the raw text to a contextualized embedding. 
Given a text sequence, we split it into tokens using the BERT tokenizer~\cite{kenton2019bert}. The BERT encoder~\cite{kenton2019bert} takes the tokens as input and outputs the contextualized embedding of each token at the output. These embeddings are denoted as $X$ ($x_1, \ldots, x_m$), where $m$ is the number of tokens in the  text sequence.

\subsection{Parameter Generation of Gaussian Mixtures} \label{sec:paramgen}
We use the parameter generator $\pi$ with the inputs $X$ and $K$ to generate the parameters $\mu_k(X), \sigma_k^2(X)$ for each component of the mixture $k \in \{1, \dots K\}$.  For simplicity, we assume a restricted form of Gaussian mixture that assigns equal probability to each Gaussian component. Further, we also assume that Gaussian components are axis-aligned that is, their covariance matrix is diagonal. Specifically, the probability distribution over the embedding space $\mathbb{R}^d$ induced by the input text embeddings $X$ is as follows:
\begin{equation}
    p_X(z) = \frac{1}{K} \sum_{k=1}^{K} \mathcal{N}(z; \mu_k(X), \sigma_k^2(X))
\end{equation}

Given an input sequence of text $X$ with token embedding representations $x_1 \dots x_{|X|}$, we initialize $K$ trainable embeddings $e_1, \ldots, e_K$ with same dimensions as $x_i$. These trainable embeddings are used to attend on $X$ to get attended token representations $a_1, \ldots, a_K$. That is, $a_k = \sum_{i=1}^m \alpha_{ik} x_i$, where $\alpha_{ik}$ are the normalized attention weights and are defined as follows:
\begin{align}
    \alpha_{ik} = \frac{\exp(x_i^\T e_k)}{\sum_{\bar{i}=1}^m \exp(x_{\bar{i}}^\T e_k)}\,, 1\le k \le K
\end{align}
Finally, the attended token embeddings are applied through two linear maps in parallel to generate the mean and log-variance of each Gaussian component in the mixture. That is, $\mu_k = f_1(a_k)$ and $\log(\sigma_k^2)) = f_2(a_k)$, where $f_1$ and $f_2$ are linear maps.

\subsection{Context and Response Encodings } Given the dialog context $c$ and response $r$, we generate the Gaussian Mixture representations $p_c(z)$ (for context) and $p_r(z)$ (for response) using $\pi$, with $K$ and $L$ components respectively. The Gaussian components of the mixture are denoted as $p_c(z; k)$ (for context) and $p_r(z; \ell)$ (for response) and are given by
\begin{equation} 
      p_c(z; k) = \mathcal{N}(z; \mu_k(\cc), \sigma_k^2(\cc)) \label{eq:context_GMM}
\end{equation}
\begin{equation}
      p_r(z; \ell) = \mathcal{N}(z; \mu_\ell(\rr), \sigma_\ell^2(\rr)) \label{eq:response_GMM}
\end{equation}
where $\mu_k(\cc)$ and $\sigma_k^2(\cc)$ are the means and variances of the $k^{th}$ Gaussian component for the context, and $\mu_\ell(\rr)$ and $\sigma_\ell(\rr)$ are the means and variances of the $\ell^{th}$ Gaussian component of the response. The parameters of the text encoders (BERT and $\pi$ module) for context and response are not shared.   

\subsection{Scoring Function}
We want the context distribution to be `close' to the distribution of the ground-truth response while simultaneously being away from distributions induced by other responses. 
We use the $\KL$ divergence to quantify this degree of closeness. The $\KL$ divergence between the distributions $p_r$ and $p_c$ over the embedding space $\mathbb{R}^d$ is given by
\begin{equation}
    \KL(p_r||p_c) = \int_{z\in \mathbb{R}^d} p_r(z)\log \frac{p_r(z)}{p_c(z)} \mathrm{d} z
\end{equation}
This integral has a closed form expression if both $p_r$ and $p_c$ are Gaussian. However, for Gaussian mixtures, this integral needs to be approximated. We derive the following approximation to the $\KL$ divergence between two GMMs.
\begin{theorem} \label{theorem1}
Let $p_r$ and $p_c$ be two Gaussian mixture distributions be $L$ and $K$ Gaussian components as defined in \eqref{eq:context_GMM} and \eqref{eq:response_GMM} respectively.
The $\KL$ divergence between the the two GMMs can be approximated by the following quantity
\begin{equation}
    \KL(p_r||p_c) \approx \frac{1}{L} \sum_{l=1}^L\min_{k \in \{1,\ldots, K\}} \KL(p_r(.;\ell)||p_c(.;k)) + \log (K/L)\,, \label{eq:approx}
\end{equation}
where $p_c(.;k)$ and $p_r(.;\ell)$ are the $k^{th}$ and $\ell^{th}$ Gaussian component of the context and response distributions as defined in~\eqref{eq:context_GMM} and \eqref{eq:response_GMM}.
\end{theorem}
A detailed derivation of the above approximation is provided in the Appendix. Note that the theorem above holds even when when the individual components of the mixture are not Gaussian. 

Intuitively, the approximation for $\KL$ divergence works as follows. For each Gaussian component in the response distribution, we find the closest Gaussian component in the context distribution. We compute the $\KL$ divergence between these neighboring components and average it over all the Gaussian components in the response.

When the components are Gaussian, the $\KL$ divergence between the components can be tractably computed using the following equation:
\begin{equation}
    \begin{aligned}
    &\KL(p_r(.;\ell)||p_c(.;k)) \\ 
    = & \frac{1}{2}\sum_{j=1}^d \left[ \log \frac{\sigma_{kj}^2(\cc)}{\sigma_{\ell j}^2(\rr)}  + \frac{\sigma_{\ell j}^2(\rr) + (\mu_{\ell j}(\rr) - \mu_{kj}(\cc))^2}{\sigma_{kj}(\cc)^2} - K\right]\,,
    \end{aligned}  \label{eq:kl_gauss}
\end{equation}
where $d$ is the dimension of the embedding space. Using equations~\eqref{eq:approx} and \eqref{eq:kl_gauss}, we get a closed form approximation to the Kullback-Leibler divergence between context and response GMMs.

\subsection{Loss Function}~\label{sec:loss_function}
We use $N$-pair contrastive loss~\cite{sohn2016improved} for training the distributions induced by the context and response. Intuitively, given a batch $\mathcal{B}$ of context-response pairs, we minimize the $\KL$ divergence between the context and the true response while simultaneously maximixing the $\KL$ divergence with respect to other randomly selected responses. The loss for a given context-response pair $(\cc, \rr)$ can be written as
\begin{equation}
    \text{loss} = \frac{\exp (-\KL(p_r||p_c))}{\sum_{\bar{r} \in \mathcal{B}} \exp(-\KL (p_{\bar{r}}|| p_c))}
\end{equation}
We average this loss across all the context-response pairs in the batch and minimize it during training. The BERT encoders, the randomly initialized embeddings as well as the linear layers for computing the means and variances, are trained in an end-to-end manner.

\subsection{Relationship with ColBERT and SBERT}
The approximation for $\KL$ divergence that we derived in equation \eqref{eq:approx}, shares a subtle relationship with the expression for similarity used in ColBERT~\cite{khattab2020colbert} and SBERT \cite{SBERT}. Let $\{c_1, \ldots, c_m\}$ and $\{r_1, \ldots, r_n\}$ be the contextualized token embeddings at the last layer of BERT for context and response respectively. The ColBERT similarity between the context and response is given by
\begin{equation}
    Sim(\cc, \rr) = \sum_{i=1}^m \max_{1\le j \le n} Sim(c_i, r_j) \,,
\end{equation}
where $Sim(c_i, r_j)$ is the inner product between the contextualized token embeddings. Thus, for each context token, ColBERT finds the most similar response token (in embedding space) and computes the similarity between these two. This similarity is then averaged over all the tokens in the response.

Instead, the $\KL$ divergence approximation derived in equation \eqref{eq:approx} finds the closest Gaussian component of the context GMM for each Gaussian component in the response GMM. Next, the $\KL$ divergence is computed between these neighboring components and averaged over all the Gaussian components in the response GMM. 

The $\KL$ divergence approximation derived in equation \eqref{eq:approx} reduces to the negative of ColBERT similarity (up to a scalar coefficient) when the following restrictions are imposed on the context and response distributions:
\begin{itemize}
    \item The Gaussian components in the context and response GMM have identity covariance. This however, makes the model less expressive. Instead, our model uses a trainable diagonal co-variance matrix.
    \item The number of Gaussian components in context and response equals the number of tokens in the context and response respectively.
    \item The means of the Gaussian components have unit norm.
\end{itemize}

Further, if we use single Gaussian mixture components, under similar assumptions as above, the model reduces to SBERT. 

\subsection{Inference}
During inference, we are provided a context and a collection of responses to select from. We map the context as well as the list of responses to their corresponding probability distributions over the embedding space.
Next, we compute the $\KL$ divergence between the distribution induced by the context and every response in the list. Using the equation derived in~\eqref{eq:approx}, this can be computed efficiently and involves standard matrix operations only. We select the top-$m$ responses that have the least $\KL$ divergence, where $m$ is specified during evaluation. 

\section{Experiments}

We answer the following questions through our experiments: (1) How does our model compare with recent dual-encoder based retrieval systems for the task of response retrieval? (2) Are the responses retrieved by our model more relevant and diverse? (3) Do human users of our system notice a difference in quality of response as compared to the recent, ColBERT system? 
\subsection{Datasets}
We conduct our experiments on two publicly available datasets -- Ubuntu Dialogue Corpus~\cite{lowe2015ubuntu}(v2.0)\footnote{https://github.com/rkadlec/ubuntu-ranking-dataset-creator} and the Twitter Customer Support Dataset\footnote{https://www.kaggle.com/thoughtvector/customer-support-on-twitter}, and one internal dataset. The Ubuntu Dialog Corpus v2.0 contains $~500K$ context-response pairs in the training set and $~20K$ context-response pairs in the validation set and test set respectively. The conversations deal with technical support for issues faced by Ubuntu users. 
The Twitter Customer Support Dataset contains $\sim 1$ million context-response pairs in the training data and $\sim 120K$ context-response pairs in validation and test sets. The conversations deal with customer support provided by several companies on Twitter.

We also conduct our experiments on an internal real-world technical support dataset with $\sim 127K$ conversations.  We will refer to this dataset as `Tech Support dataset' in the rest of the paper. The Tech Support dataset contains conversations pertaining to an employee seeking assistance from an agent (technical support) --- to resolve problems such as password reset, software installation/licensing, and wireless access. In contrast to Ubuntu dataset, which used user forums to construct the data,  this dataset has clearly two distinct users --- employee and agent. In all our experiments we model the \textit{agent} response turns only.

For each conversation in the Tech Support dataset, we sample context and response pairs. 
Note that multiple context-response pairs can be generated from a single conversation. For each conversation, we sample $25\%$ of the possible context-response pairs. We create validation pairs by selecting $5000$ conversations randomly and sampling their context response pairs. Similarly, we create test pairs from a different subset of $5000$ conversations. The remaining conversations are used to create training context-response pairs. 

\subsection{Baselines}
We compare our proposed model against two scalable baselines - SBERT~\cite{SBERT} and ColBERT~\cite{khattab2020colbert} --  a recent state-of-the-art retrieval model. Similar to Mix-and-Match, both the baselines use independent encoders (dual-encoders to encode the contexts and responses. Hence, these baselines can be used for large-scale retrieval at an acceptable cost.
\subsubsection{SBERT} SBERT~\cite{SBERT} uses two BERT encoders for embedding the inputs (context and response). The contextualized embeddings at the last layer are pooled to generate fixed size embeddings for context and response. Since context and response are from two different domains, we force the two BERT encoders to not share the parameters. We use inner product between the context and response embeddings as the similarity measure and train the two encoders via contrastive loss.
\subsubsection{ColBERT} Just like SBERT, ColBERT~\cite{khattab2020colbert} uses two BERT encoders to encode the inputs. However, instead of pooling the contextualized embeddings at the last layer, a late interaction is computed between all the contextualized token embeddings of the context and response. Unlike the original implementation of ColBERT, we do not enforce the context and response encoders to share parameters. This is essential for achieving reasonable performance for dialogs. The model is trained via contrastive loss.

\subsection{Model and training details}
We use the pretrained `bert-base' model provided by Hugging Face\footnote{https://huggingface.co/bert-base-uncased}. The dimension of the embedding space is fixed to be $128$ for all the models. The number of Gaussian components in the context and response distributions is selected by cross-validation from the set $\{1,2,4,8,16,32, \text{all}\}$. Here, the `all' setting refers to the case where the context/response distribution has as many Gaussian components as the number of tokens in the context/response. We use the `AdamW' optimizer provided by Hugging Face (Adam optimizer with a fixed weight decay) with a learning rate of $1.5e-5$ for all our experiments. A fixed batch size of $16$ context-response pairs is used. To prevent overfitting, we use early-stopping with the loss function defined in Section~\ref{sec:loss_function} on validation set as the stopping criteria.

\begin{table*}
\begin{center}
\caption{Comparison of Mix-and-Match against baselines on retrieval tasks. Given a context, the task involves retrieving from a set of $5000$ responses that also contains the ground truth response.}
\begin{tabular}{|c|c|c|c|c|c|}
\hline
{\bf Dataset} & {\bf Model} & {\bf Recall@2} & {\bf Recall@5} & {\bf Recall@10} & {\bf MRR}
\\
\hline
& SBERT & 8.44 & 13.26 & 18.26 & 0.099 
\\
Ubuntu (v2) & ColBERT & 10.93 & 16.37 & 21.33 & 0.123
\\
& \textbf{Mix-and-Match} & \textbf{17.44} & \textbf{24.27} & \textbf{29.75} & \textbf{0.167}
\\
\hline
& SBERT & 9.82 & 19.08 & 29.64 & 0.135
\\
Twitter & ColBERT & 12.62 & 20.36 & 34.82 & 0.137
\\
& \textbf{Mix-and-Match} & \textbf{16.06} & \textbf{28.58} & \textbf{40.54} &
\textbf{0.195} \\
\hline
& SBERT & 7.71 & 12.69 & 22.67 & 0.119
\\
Tech Support & ColBERT & 8.82 & 14.97 & 23.91 & 0.125
\\
& \textbf{Mix-and-Match} & \textbf{9.67} & \textbf{15.68} & \textbf{26.47} & \textbf{0.133}
\\
\hline
\end{tabular}
\vspace{3mm}

\label{table:recall}
\end{center}
\end{table*}

\subsection{Response Retrieval}
In this setting, each context is paired with $5000$ randomly selected responses along with the ground truth response for the given context. The list of $5000$ responses are randomly selected from the test data for each instance. Hence, the response universe associated with each dialog-context may be different. The task then is to retrieve the ground truth response given the context. For efficient computation, the full universe of responses are encoded once and stored. Note that is only possible for dual-encoder architectures (such as Mix-and-Match, SBERT, ColBERT); the major performance bottleneck in cross-encoder approaches arises from this step where the response encodings are dependent on the context and hence need to be encoded each time for every new dialog context. 

For Mix-and-Match, the response encoder outputs the means and variances of the GMM induced by the response in the embedding space. We use a batch-size of $50$ to encode the responses and cache the generated parameters (mean and variance) of the response-GMMs. 

Similarly, the context is encoded by the context encoder to output the means and variances of the components of context-GMM. We compute the $\KL$ divergence between the context distribution and distribution of each response in the associated list of $5000$ responses using the expressions derived in \eqref{eq:approx} and \eqref{eq:kl_gauss}. The values are sorted in ascending order and the top-$k$ responses are selected for evaluation.

A similar setting is used for SBERT and ColBERT with the exception that the embeddings are stored instead of means and variances. Moreover, we sort the responses based on SBERT and ColBERT similarity in descending order.

\subsubsection{Results}
We use MRR and Recall@k for evaluating the various models. For evaluating MRR, we sort the associated set of $5000$ responses with each context, based on $\KL$ divergence in ascending order. Next, we compute the rank of the ground truth response in the sorted list. The MRR is then obtained as the mean of the reciprocal rank for all the contexts. For Recall@k, we pick the top-$k$ responses with the least $\KL$ divergence. The percentage of contexts for which the ground truth response is present in the top-$k$ responses is referred to as Recall@k. The results are shown in Table~\ref{table:recall}. 

As can be observed, SBERT that uses a single embedding to represent the entire context as well as response, achieves the lowest recall. By using all the token embeddings to represent the context and response, ColBERT achieves better performance than SBERT. Finally, by using Gaussian mixture probability distributions to represent context and response, Mix-and-Match achieves substantial improvement in recall@k and MRR on all the datasets as compared to SBERT and ColBERT. Thus, richer the representation of context and response, better is the recall. Note that the relative improvement is less in Tech Support as there is less diversity among the responses in the training data of Tech Support. The agents are trained to handle calls in specific way that reduces the diversity.

\subsection{Response Recommendation} \label{sec:resp_rec}
The response retrieval setting described in the previous section is unrealistic since it assumes that the ground truth response is also present in a set of $5000$ responses. In reality, when a response retrieval model such as~\cite{AgentAssist} is deployed for response recommendation, it must retrieve from a large set of all the responses present in the training data (often running into hundreds of thousands of responses).  

To deal with the large set of responses present in the training data, we encode them offline using the response encoder of Mix-and-Match. As in the previous section, we use a batch-size of $50$ for encoding the responses. After the means and variances of all the Gaussian components of response GMMs have been generated, we save them to a file along with the corresponding responses. To ensure faster retrieval, we use \textrm{Faiss}~\cite{faiss} for indexing the means of the Gaussian components of response GMMs. Faiss is a library for computing fast vector-similarities and has been used for vector-based searching in huge sets. We use the IVFPQ index of faiss (Inverted File with Product Quantization) that discretizes the embedding space into a finite number of cells. This allows for faster search computations.

We flatten the tensor of means of Gaussian components of all response GMMs to a matrix of mean vectors. The matrix of mean vectors is added to the IVFPQ index. A pointer is maintained from the mean of each Gaussian component to the corresponding response as well as the means and variances of its Gaussian components. 

When a new context arrives, we compute the means and variances of its Gaussian components. For each Gaussian component, we retrieve the top-10 responses by using the mean of the Gaussian component as the search query. After retrieving the top-10 responses for each Gaussian component, we load the corresponding means and variance. Finally, we compute the $\KL$ divergence between the context GMM and the GMMs of all the retrieved responses. The values are sorted in ascending order and the top-$k$ responses are selected for evaluation. 

\begin{table}
\begin{center}
\small
\caption{Comparison of Mix-and-Match against baselines for the response recommendation task. Given a context, the task involves retrieving from the set of all responses in the training data. To handle the large set of responses, we use a FAISS~\cite{faiss} index for pre-retrieval. The computation of diversity are discussed in detail in Section~\ref{sec:resp_rec} }
\begin{tabular}{|c|c|c|c|c|}
\hline
{\bf Dataset} & {\bf Model} & {\bf BLEU-2} & {\bf BLEU-4} & \begin{tabular}[c]{@{}c@{}}\textbf{Diversity }\\\textbf{(BERTDist.)}\end{tabular}  
\\
\hline
& SBERT & 5.86 & 0.49 & 2.33
\\
Ubuntu (v2) & ColBERT & 6.66 & 0.58 & 3.19
\\
& \textbf{Mix-and-Match} & \textbf{7.16} & \textbf{0.64} & \textbf{3.60}
\\
\hline
& SBERT & 19.84 & 10.3 & 1.76
\\
Twitter & ColBERT & 20.67 & 11.09 & 2.17
\\
& \textbf{Mix-and-Match} & \textbf{22.83} & \textbf{12.62} & \textbf{2.60}
\\
\hline
& SBERT & 12.09 & 5.82 & 1.49
\\
Tech Support & ColBERT & 16.57 & 8.58 & 2.55
\\
& \textbf{Mix-and-Match} & \textbf{18.82} & \textbf{10.57} & \textbf{3.02}
\\
\hline
\end{tabular}
\vspace{3mm}

\label{table:bleu}
\end{center}
\end{table}

\subsubsection{BLEU}
Since the ground truth response may not be present verbatim in the set, metrics such as recall and MRR cannot be computed in this setting. 
We therefore use the BLEU metric~\cite{papineni2002bleu} for evaluating the quality of the responses. The BLEU metric measures the count of ngrams that are common between the ground truth and predicted response. The results are shown in Table~\ref{table:bleu}. As can be observed from the table, the BLEU scores are quite low for Ubuntu dataset, suggesting that most retrieved responses have very little overlap with the ground truth response. As in the previous section, SBERT is outperformed by ColBERT in terms of BLEU-2 and BLEU-4. Finally, Mix-and-Match outperforms both the models on all three datasets. This suggests that the responses retrieved by Mix-and-Match are relevant to the dialog context.

\subsubsection{Diversity}
The primary strength of the Mix-and-Match system is its capability to associate multiple diverse responses with the same context. To capture the diversity among the top-$k$ responses retrieved for a given context, we measure the distance between every pair of responses and average it across all pairs. Thus, if $\mathcal{R}$ is the set of retrieved responses for a given context, the BERT distance among the responses in $\mathcal{R}$ is given by
\begin{equation}
    \mathrm{BERT Distance}(\mathcal{R}) = \frac{1}{\mathcal{|R|}^2} \sum_{\rr \in \mathcal{R}}\sum_{\bar{\rr} \in \mathcal{R}} ||e(\rr) - e(\bar{\rr})||^2\,,
\end{equation}
where $e(\bar{\rr})$ is the pooled BERT embedding of $\rr$. 

The results are shown in Table~\ref{table:bleu}. As can be observed from the table, SBERT has the least diversity among the retrieved responses. This is expected since all the retrieved responses must be close to the context embedding and hence, close to each other. ColBERT fares better in terms of diversity since it uses multiple embeddings to represent contexts and responses. Finally, Mix-and-Match that uses GMMs to represent contexts and responses achieves the best diversity. This suggests that having multiple or probabilistic embeddings helps in improving the diversity among the retrieved responses.

\subsubsection{Scalability}
Next, we evaluate the time taken by the Mix-and-Match model to retrieve from the FAISS index as comapred to baselines. The similarity/$\KL$-divergence computations as well as vector similarity searches for the FAISS index, are performed on a single A100 GPU. Unsurprisingly, SBERT achieves the lowest latency of $8.9$ ms for retrieval per dialog context. ColBERT achieves a latency of $89.7$ ms. The latency of Mix-and-Match ranges from $36.7$ ms to $477.8$ ms depending upon the number of Gaussian components in the mixture. Note that, even in the worst case, the latency is less than $0.5$s, thus making the model suitable for practical use in the real world.

SBERT, ColBERT and Mix-and-Match use independent encoders to encode the responses. Hence,  response encoding can be done offline. During inference, the context is encoded once and its similarity /$\KL$ divergence with the pre-encoded responses is computed. In contrast, for models that use joint encoding~\cite{Wu2017SequentialMN, BERT}, the context must be jointly encoded with every response during inference. Thus, the time taken by joint encoding approaches is proportional to the number of responses in the retrieval set, making these approaches unsuitable for practical real-world deployment. 

\begin{table}
\centering
\small
\caption{The top-response returned by the Mix-and-match model is found to be relevant more often (40\% vs 17\%) than ColBERT. In addition, the set of responses returned by Mix-and-Match are also more diverse (58\% vs 42\% for ColBERT).}
\begin{tabular}{|c|c|c|c|} 
\hline
\multicolumn{1}{|l|}{}                                                                & \textbf{ColBERT Win}  & \textbf{Mix and Match Win} & \textbf{Tie}  \\ 
\hline
\begin{tabular}[c]{@{}c@{}}\textbf{Response }\\\textbf{Relevance @1}\end{tabular}     & \multicolumn{1}{c|}{17\%   } &          \textbf{40\%}               &      43\%         \\ 
\hline
\cline{1-1}
\textbf{Diversity}           &      42\%                                                         &                  \textbf{58\%}     &           NA                             \\
\hline
\end{tabular}

\label{tab:human}
\end{table}
\begin{table}
\centering
\small
\caption{The Diversified-Relevance scores for ColBERT and Mix-and-Match in our human study.}
\begin{tabular}{|l|l|l|} 
\hline
                                                 & \textbf{ColBERT }         & \textbf{Mix and Match~}  \\ 
\hline
\multicolumn{1}{|c|}{Diversified-Relevance (DR)} & \multicolumn{1}{c|}{0.25} & \multicolumn{1}{c|}{{\bf 0.35}}                    \\
\hline
\end{tabular}
\label{tab:DR-scores}
\end{table}
\begin{table*}[ht]
\centering
\small
\caption{Sample of a single-turn dialog context - Mix-and-Match returns a relevant response at the top ranked position and another related response at the second position. In contrast, ColBERT retrieved generic or unrelated responses.}
\begin{tabular}{l|l} 
\hline
\multicolumn{2}{c}{\textbf{Dialog Context}}                                                                                                                                                                                                                                                                                                                                                                                                                                                                                                                                                                                                                                                                                                                       \\ 
\hline
\multicolumn{2}{l}{\textbf{User:}\textit{ @southwestair i'm bummed i paid 15 for early bird and still only got a b boarding position don t think i ll do that anymore notworthit}}                                                                                                                                                                                                                                                                                                                                                                                                                                                                                                                                                                                \\ 
\hline
\multicolumn{2}{c}{\textbf{Responses Retrieved}}                                                                                                                                                                                                                                                                                                                                                                                                                                                                                                                                                                                                                                                                                                                  \\ 
\hline
\multicolumn{1}{c|}{\textbf{ColBERT}}                                                                                                                                                                                                                      & \multicolumn{1}{c}{\textbf{Mix-and-Match}}                                                                                                                                                                                                                                                                                                                                                                                                                                                           \\ 
\hline
\begin{tabular}[c]{@{}l@{}}(i)if you check your itinerary email, it ' ll tell you what type of \\fare you purchased.wanna get away fares are nonrefundable, \\but anytime and business select fares can be refunded.\\thanks for reaching out\end{tabular} & \begin{tabular}[c]{@{}l@{}}(\textcolor[rgb]{0,0.502,0}{\textbf{i) please know that earlybird offers the convenience }}\\\textcolor[rgb]{0,0.502,0}{\textbf{of automatic~ check - in, but does not guarantee a }}\\\textcolor[rgb]{0,0.502,0}{\textbf{specific boarding position . still, we apologize if you}}\\\textcolor[rgb]{0,0.502,0}{\textbf{~were disappointed in the position you received \&hope to }}\\\textcolor[rgb]{0,0.502,0}{\textbf{make it up to~ you while onboard}}\end{tabular}  \\ 
\hline
\begin{tabular}[c]{@{}l@{}}(ii) sorry if we've missed a chance to be of better service to you.\\since we're separate entities,~ we're re unable to alter or change\\the incorrect alamo car reservation\end{tabular}                                        & \begin{tabular}[c]{@{}l@{}}(ii) by law we ' re required to let folks with specific boarding needs\\~or disabilities on the aircraft first, though we apologize for your\\~frustration this morning\end{tabular}                                                                                                                                                                                                                                                                                      \\ 
\hline
\begin{tabular}[c]{@{}l@{}}(iii) oh no ! so sorry to hear that. please speak with our agents in the \\airport about reaccommodations\end{tabular}                                                                                                          & \begin{tabular}[c]{@{}l@{}}(iii) sorry for any confusion, our agents know the proper \\procedures and questions toask to determine the best boarding option\end{tabular}                                                                                                                                                                                                                                                                                                                             \\ 
\hline
\multicolumn{2}{l}{\begin{tabular}[c]{@{}l@{}}\textbf{Ground Truth Response:} apologies for any frustration, as the \# of earlybird customers vary on each flt, you're guaranteed automatic check in, \\not a specific boarding position .\end{tabular}}                                                                                                                                                                                                                                                                                                                                                                                                                                                                                                          \\
\hline
\end{tabular}

\label{tab:qual1}
\end{table*}

\begin{table*}[ht]
\centering
\small
\caption{Sample of a multi-turn dialog context - Mix-and-Match returns a relevant response at the top ranked position and related responses at other positions. In contrast, ColBERT retrieved generic or unrelated responses.}
\begin{tabular}{l|l} 
\hline
\multicolumn{2}{c}{\textbf{Dialog Context}}                                                                                                                                                                                                                                                                                                                                                                                                         \\ 
\hline
\multicolumn{2}{l}{\begin{tabular}[c]{@{}l@{}}\textbf{User:} the worst mobile serive in 2015 2017 cellphone badservice miami florida\\\textbf{Agent:} hey send us a dm and we'll ensure a great experience channeyt\end{tabular}}                                                                                                                                                                                                                   \\
\multicolumn{2}{l}{\textbf{User:}\textit{ tmobilehelp poor service low signal slow service it s miami}}                                                                                                                                                                                                                                                                                                                                             \\ 
\hline
\multicolumn{2}{c}{\textbf{Responses Retrieved}}                                                                                                                                                                                                                                                                                                                                                                                                    \\ 
\hline
\multicolumn{1}{c|}{\textbf{ColBERT}}                                                                                                                                                               & \multicolumn{1}{c}{\textbf{Mix-and-Match}}                                                                                                                                                                                                    \\ 
\hline
\begin{tabular}[c]{@{}l@{}}(i) Our apologies , we are currently experiencing a system challenge \\which~ we are working to resolve . kindly bear with us.\end{tabular}                              & \begin{tabular}[c]{@{}l@{}}\textbf{\textcolor[rgb]{0,0.392,0}{(i) how long has this been happening ? what type of phone}}\\\textbf{\textcolor[rgb]{0,0.392,0}{do you have ?~ please send us a dm so we can fix it . thank you}}\end{tabular}  \\ 
\hline
\begin{tabular}[c]{@{}l@{}}(ii) our sincere apologies for any inconveniences caused, we are\\having a~ technical issue,~ resolution is underway\end{tabular}                                        & \begin{tabular}[c]{@{}l@{}}(ii) that ' s not good at all ! please dm us with your zip code and \\nearest streets intersection to check the coverage\end{tabular}                                                                              \\ 
\hline
\begin{tabular}[c]{@{}l@{}}(iii) it is not our intention to make you upset. please feel free\\~to reach out to us if you have already called back and still need further\\~assistance.\end{tabular} & \begin{tabular}[c]{@{}l@{}}(iii) does this happen in specific locations ? when did you begin to\\experience these issues with your connection ? are you having issues \\making calls and sending text as well ?\end{tabular}                  \\ 
\hline
\multicolumn{2}{l}{\textbf{Ground Truth Response}: let ' s flip thing around ! meet in the dms https://t.co/sbivwmm6x2}                                                                                                                                                                                                                                                                                                                             \\ 
\hline
\multicolumn{2}{l}{}                                                                                                                                                                                                                                                                                                                                                                                                                               
\end{tabular}

\label{tab:qual2}
\end{table*}
\subsection{Human evaluation}
We also conducted a human study comparing the output responses of ColBERT and Mix-and-match. We used samples from the Twitter data set for this study as it does not require domain expertise to assess the relevance of responses. Three users were asked to review $30$ twitter dialogs contexts along with the top-4 responses returned by each system,\footnote{a total of $360$ independent context-response assessments.} in a response recommendation setting. Users were presented the outputs from each system in random order and they were blind to the system returning the responses. We asked our users the following: 
\begin{enumerate}
    \item Given the dialog context and the response sets from two different systems, label each response with a ``yes'' or ``no' depending on whether the response is a relevant response recommendation for the dialog context. Thus, each response returned by both systems was individually labeled by three human users.
    \item Given the dialog context and the response sets from two different systems, which of the response set is more diverse? Thus, each context-recommendation set was assessed by three human users.
\end{enumerate}

We count the number of votes received by the top-ranked response for each system and report percentage wins for each system. In addition, we also report a head-to-head comparison in which the two models were assessed for diversity (no ties). Finally, to assess whether diversity is accompanied by relevance in the response set, we define a metric called {\it Diversified-Relevance (DR)} which weighs the diversity wins by the number of relevant responses returned by each system. Specifically, $DR^{model}$, the Diversified-Relevance for a $model \in \{$ColBERT, Mix-and-Match$\}$ is given by:
\begin{equation}
   DR^{model} =  \frac{\sum_i^M \sum_j^4\mathbbm{1\{}win^{model}_i\} * \mathbbm{1}\{relevance^{model}_{ij}\}}{4M},
\end{equation}
where $M$ is the number of dialogs used in the human study, $4$, is the number of response recommendations per dialog,  $\mathbbm{1}\{win^{model}_i\}$ is an indicator function that takes the value $1$ if $model$ was voted as being more diverse its responses to $i^{th}$ dialog context, and $\mathbbm{1}\{relevance^{model}_{ij}\}$, is an indicator function that takes the value $1$ if the $j^{th}$ response recommendation by $model$ was voted as being relevant\footnote{As can be seen $DR$ returns a score between 0 and 1. }. 


\subsubsection{Results }As can be seen in Table \ref{tab:human}, the top-ranked response returned by Mix-and-Match received significantly higher number of votes (40\%) in favour as compared to ColBERT. In 43\% of the cases there was no-clear winner. Finally, in 58\% of the dialogs, Mix-and-Match was found to present a more diverse set of response recommendations.

In order to assess, if the diversity is accompanied by relevance, we also report the $DR$ scores in Table \ref{tab:DR-scores}. As can be seen the DR scores for Mix-and-Match is significantly higher than ColBERT (0.35 vs 0.25). Overall, the results from our human-study indicate that Mix-and-Match returns more diverse and relevant responses.

\subsection{Qualitative Study}

We present two sample outputs in Tables \ref{tab:qual1} and Table \ref{tab:qual2}; Table \ref{tab:qual1} shows a sample with a single-turn dialog context where the user is complaining about flight boarding positions. The responses retrieved by both ColBERT and Mix-and-Match are presented. As can be seen, Mix-and-Match returns a relevant response at the top ranked position (highlighted in \textcolor[rgb]{0,0.392,0}{\textbf{green}}) and another related response at the second position. In contrast, ColBERT retrieved generic or unrelated responses.

Table \ref{tab:qual2} shows a sample with a multi-turn dialog context where the user is complaining about bad cellphone coverage. As before, the responses retrieved by both ColBERT and Mix-and-Match are presented. As can be seen, Mix-and-Match returns a relevant response at the top ranked position (highlighted in \textcolor[rgb]{0,0.392,0}{\textbf{green}}) and related responses at other positions. In contrast, ColBERT retrieved generic or unrelated responses.

\vspace{-1ex}
\section{Conclusion}
In this paper we presented a dialog response retrieval method called  - Mix-and-Match, which is designed to accommodate the many-to-many relationships that exist between a dialog context and responses. We modeled the dialog context and response using mixtures of gaussians, instead of point embeddings. This allows the network to be more expressive and it does not force the representations of unrelated responses to move closer, as would have been the case with traditional dual-encoder learning objectives. We derived and presented a closed form expressions for efficiently computing the KL-divergence based distance measures and showed its suitability for real-world settings. We also related our model to existing retrieval methods, SBERT and ColBERT, under specific assumptions about the nature of the GMMs. We demonstrated the effectiveness of our retrieval systems on three different datasets - Ubuntu, Twitter and an internal, real-world Tech support dataset. Additional experiments for response relevance,including a human study were performed on the publicly available datasets. We found that not only is our model able to retrieve more relevant responses as compared to recent retrieval systems, it also presented more diverse results. This is especially important for response recommendation systems \cite{AgentAssist} where human agents may chose from a set of recommendations.
\vspace{-4ex}
\section{Appendix}
\subsection{Proof of Theorem~\ref{theorem1}}
\begin{proof}
The proof follows a similar line of reasoning as the proof provided in~\cite{kl_gmm_approx}
The $\KL$ divergence between $p_r$ and $p_c$ can be written as
\begin{align}
    KL(p_r|| p_c) &= \int p_r(z)\log p_r(z) \dz - \int p_r(z)\log p_c(z)\dz \notag\\
                &= -\HH(p_r) + \HH(p_r, p_c) \label{eq:KL_exp}
\end{align}
 The first term is the negative of entropy while the second term is the cross entropy. We approximate the cross entropy by expanding the GMM in terms of its Gaussian components, and applying Jensen's inequality:
\begin{align}
    \HH &(p_r, p_c) \notag \\
    & = -\frac{1}{L}\sum_{\ell=1}^L\int p_r(z; \ell) \log \left[ \sum_{k=1}^K q_\ell(k) \frac{p_c(z; k)}{q_\ell(k)K}\right]\dz \notag \\
    & \le -\frac{1}{L}\sum_{\ell=1}^L \sum_{k=1}^K q_\ell(k) \int p_r(z; \ell) \log  p_c(z; k)\dz \notag \\
    & \hspace{1cm} \frac{1}{L}\sum_{\ell=1}^L \sum_{k=1}^K q_\ell(k) \log q_\ell(k) + \log K \notag \\
    & = \frac{1}{L}\sum_{\ell=1}^L \sum_{k=1}^K q_\ell(k)  \HH(p_r(.;\ell), p_c(.;k)) - \HH(q_\ell) + \log K
\end{align}
Here, the first equality follows by multiplying and dividing the terms within the $\log$ by the variational distribution $q_\ell(k)$. The last inequality follows by applying Jensen's inequality. The above upper bound holds for all choice of $q$. The bound can be tightened by minimizing it with respect to $q_\ell(k)$. We assume $q_\ell$ to be a one-hot vector which can only be non-zero for one context component $k$. Every one-hot $q_\ell$ has an entropy of $0$ and hence, the second term in the equation is always $0$.
For a one-hot $q_\ell$, the above equation is minimized when $q_\ell$ assigns all its weights to the component of context GMM with lowest cross-entropy. Using the optimal one-hot $q$, the above equation can be written as
\begin{align}
    \HH (p_r, p_c) \le \frac{1}{L}\sum_{\ell=1}^L \min_{k \in \{1, \ldots, K\}} \HH(p_r(.;\ell), p_c(.;k)) + \log K \label{eq:cross_approx}
\end{align}

The entropy of $p_r$ can be derived as a special case of the above equation by replacing $p_c$ in the above equation by $p_r$. Thus, the entropy of a GMM can be upper-bounded by
\begin{align}
    \HH (p_r) \le \frac{1}{L}\sum_{\ell=1}^L  \HH(p_r(.;\ell)) + \log L \label{eq:ent_approx}
\end{align}
Finally, the $\KL$ divergence can be approximated by replacing \eqref{eq:cross_approx} and \eqref{eq:ent_approx} in \eqref{eq:KL_exp}. Note that the resultant quantity is neither an upper nor a lower bound, but still a useful approximation.
\begin{align}
    \KL(p_r||p_c) &\approx \frac{1}{L}\sum_{\ell=1}^L  \min_{k \in \{1, \ldots, K\}} \left[- \HH(p_r(.;\ell)) + \HH(p_r(.;\ell), p_c(.;k))\right] \\
    & \hspace{1cm}+ \log (K/L) \\
    &= \frac{1}{L}\sum_{\ell=1}^L \min_{k \in \{1, \ldots, K\}} \KL(p_r(.;\ell)|| p_c(.;k)) + \log (K/L)
\end{align}
\end{proof}


\bibliographystyle{ACM-Reference-Format}
\bibliography{bibfile}


\begin{thebibliography}{53}


\ifx \showCODEN    \undefined \def \showCODEN     #1{\unskip}     \fi
\ifx \showDOI      \undefined \def \showDOI       #1{#1}\fi
\ifx \showISBNx    \undefined \def \showISBNx     #1{\unskip}     \fi
\ifx \showISBNxiii \undefined \def \showISBNxiii  #1{\unskip}     \fi
\ifx \showISSN     \undefined \def \showISSN      #1{\unskip}     \fi
\ifx \showLCCN     \undefined \def \showLCCN      #1{\unskip}     \fi
\ifx \shownote     \undefined \def \shownote      #1{#1}          \fi
\ifx \showarticletitle \undefined \def \showarticletitle #1{#1}   \fi
\ifx \showURL      \undefined \def \showURL       {\relax}        \fi
\providecommand\bibfield[2]{#2}
\providecommand\bibinfo[2]{#2}
\providecommand\natexlab[1]{#1}
\providecommand\showeprint[2][]{arXiv:#2}

\bibitem[\protect\citeauthoryear{Aronsson, Lu, Str{\"u}ber, and
  Berger}{Aronsson et~al\mbox{.}}{2021}]%
        {dialogframeworks}
\bibfield{author}{\bibinfo{person}{Johan Aronsson}, \bibinfo{person}{Philip
  Lu}, \bibinfo{person}{Daniel Str{\"u}ber}, {and} \bibinfo{person}{Thorsten
  Berger}.} \bibinfo{year}{2021}\natexlab{}.
\newblock \showarticletitle{A maturity assessment framework for conversational
  AI development platforms}. In \bibinfo{booktitle}{\emph{Proceedings of the
  36th Annual ACM Symposium on Applied Computing}}.
  \bibinfo{pages}{1736--1745}.
\newblock


\bibitem[\protect\citeauthoryear{Athiwaratkun and Wilson}{Athiwaratkun and
  Wilson}{2017}]%
        {athiwaratkun-wilson-2017-multimodal}
\bibfield{author}{\bibinfo{person}{Ben Athiwaratkun} {and}
  \bibinfo{person}{Andrew Wilson}.} \bibinfo{year}{2017}\natexlab{}.
\newblock \showarticletitle{Multimodal Word Distributions}. In
  \bibinfo{booktitle}{\emph{Proceedings of the 55th Annual Meeting of the
  Association for Computational Linguistics (Volume 1: Long Papers)}}.
  \bibinfo{publisher}{Association for Computational Linguistics},
  \bibinfo{address}{Vancouver, Canada}, \bibinfo{pages}{1645--1656}.
\newblock
\urldef\tempurl%
\url{https://doi.org/10.18653/v1/P17-1151}
\showDOI{\tempurl}


\bibitem[\protect\citeauthoryear{Athiwaratkun, Wilson, and
  Anandkumar}{Athiwaratkun et~al\mbox{.}}{2018}]%
        {GaussianEmb}
\bibfield{author}{\bibinfo{person}{Ben Athiwaratkun}, \bibinfo{person}{Andrew
  Wilson}, {and} \bibinfo{person}{Anima Anandkumar}.}
  \bibinfo{year}{2018}\natexlab{}.
\newblock \showarticletitle{Probabilistic {F}ast{T}ext for Multi-Sense Word
  Embeddings}. In \bibinfo{booktitle}{\emph{Proceedings of the 56th Annual
  Meeting of the Association for Computational Linguistics (Volume 1: Long
  Papers)}}. \bibinfo{publisher}{Association for Computational Linguistics},
  \bibinfo{address}{Melbourne, Australia}, \bibinfo{pages}{1--11}.
\newblock
\urldef\tempurl%
\url{https://doi.org/10.18653/v1/P18-1001}
\showDOI{\tempurl}


\bibitem[\protect\citeauthoryear{Bartl and Spanakis}{Bartl and
  Spanakis}{2017}]%
        {Bartl2017ARD}
\bibfield{author}{\bibinfo{person}{Alexander Bartl} {and}
  \bibinfo{person}{Gerasimos Spanakis}.} \bibinfo{year}{2017}\natexlab{}.
\newblock \showarticletitle{A retrieval-based dialogue system utilizing
  utterance and context embeddings}.
\newblock \bibinfo{journal}{\emph{CoRR}}  \bibinfo{volume}{abs/1710.05780}
  (\bibinfo{year}{2017}).
\newblock


\bibitem[\protect\citeauthoryear{Bromley, Bentz, Bottou, Guyon, LeCun, Moore,
  S{\"a}ckinger, and Shah}{Bromley et~al\mbox{.}}{1993}]%
        {bromley1993signature}
\bibfield{author}{\bibinfo{person}{Jane Bromley}, \bibinfo{person}{James~W
  Bentz}, \bibinfo{person}{L{\'e}on Bottou}, \bibinfo{person}{Isabelle Guyon},
  \bibinfo{person}{Yann LeCun}, \bibinfo{person}{Cliff Moore},
  \bibinfo{person}{Eduard S{\"a}ckinger}, {and} \bibinfo{person}{Roopak Shah}.}
  \bibinfo{year}{1993}\natexlab{}.
\newblock \showarticletitle{Signature verification using a “siamese” time
  delay neural network}.
\newblock \bibinfo{journal}{\emph{International Journal of Pattern Recognition
  and Artificial Intelligence}} \bibinfo{volume}{7}, \bibinfo{number}{04}
  (\bibinfo{year}{1993}), \bibinfo{pages}{669--688}.
\newblock


\bibitem[\protect\citeauthoryear{Cao and Clark}{Cao and Clark}{2017}]%
        {cao2017latent}
\bibfield{author}{\bibinfo{person}{Kris Cao} {and} \bibinfo{person}{Stephen
  Clark}.} \bibinfo{year}{2017}\natexlab{}.
\newblock \showarticletitle{Latent Variable Dialogue Models and their
  Diversity}. In \bibinfo{booktitle}{\emph{Proceedings of the 15th Conference
  of the European Chapter of the Association for Computational Linguistics}},
  Vol.~\bibinfo{volume}{2}. \bibinfo{pages}{182--187}.
\newblock


\bibitem[\protect\citeauthoryear{Chen, Lv, Yi, and Yi}{Chen
  et~al\mbox{.}}{2021b}]%
        {facialreco}
\bibfield{author}{\bibinfo{person}{Kai Chen}, \bibinfo{person}{Qi Lv},
  \bibinfo{person}{Taihe Yi}, {and} \bibinfo{person}{Zhengming Yi}.}
  \bibinfo{year}{2021}\natexlab{b}.
\newblock \showarticletitle{Reliable Probabilistic Face Embeddings in the
  Wild}.
\newblock \bibinfo{journal}{\emph{CoRR}}  \bibinfo{volume}{abs/2102.04075}
  (\bibinfo{year}{2021}).
\newblock
\showeprint[arXiv]{2102.04075}
\urldef\tempurl%
\url{https://arxiv.org/abs/2102.04075}
\showURL{%
\tempurl}


\bibitem[\protect\citeauthoryear{Chen, Hui, He, Han, Sun, and Ye}{Chen
  et~al\mbox{.}}{2021a}]%
        {CoBERT}
\bibfield{author}{\bibinfo{person}{Xiaoyang Chen}, \bibinfo{person}{Kai Hui},
  \bibinfo{person}{Ben He}, \bibinfo{person}{Xianpei Han}, \bibinfo{person}{Le
  Sun}, {and} \bibinfo{person}{Zheng Ye}.} \bibinfo{year}{2021}\natexlab{a}.
\newblock \showarticletitle{Co-BERT: A Context-Aware BERT Retrieval Model
  Incorporating Local and Query-specific Context}.
\newblock \bibinfo{journal}{\emph{arXiv preprint arXiv:2104.08523}}
  (\bibinfo{year}{2021}).
\newblock


\bibitem[\protect\citeauthoryear{Chun, Oh, de~Rezende, Kalantidis, and
  Larlus}{Chun et~al\mbox{.}}{2021}]%
        {Chun2021ProbabilisticEF}
\bibfield{author}{\bibinfo{person}{Sanghyuk Chun}, \bibinfo{person}{Seong~Joon
  Oh}, \bibinfo{person}{Rafael~Sampaio de Rezende}, \bibinfo{person}{Yannis
  Kalantidis}, {and} \bibinfo{person}{Diane Larlus}.}
  \bibinfo{year}{2021}\natexlab{}.
\newblock \showarticletitle{Probabilistic Embeddings for Cross-Modal
  Retrieval}.
\newblock \bibinfo{journal}{\emph{2021 IEEE/CVF Conference on Computer Vision
  and Pattern Recognition (CVPR)}} (\bibinfo{year}{2021}),
  \bibinfo{pages}{8411--8420}.
\newblock


\bibitem[\protect\citeauthoryear{Contractor, Singla, and {Mausam}}{Contractor
  et~al\mbox{.}}{2016}]%
        {entcmp}
\bibfield{author}{\bibinfo{person}{Danish Contractor}, \bibinfo{person}{Parag
  Singla}, {and} \bibinfo{person}{{Mausam}}.} \bibinfo{year}{2016}\natexlab{}.
\newblock \showarticletitle{Entity-balanced {G}aussian p{LSA} for Automated
  Comparison}. In \bibinfo{booktitle}{\emph{Proceedings of the 2016 Conference
  of the North {A}merican Chapter of the Association for Computational
  Linguistics: Human Language Technologies}}. \bibinfo{publisher}{Association
  for Computational Linguistics}, \bibinfo{address}{San Diego, California},
  \bibinfo{pages}{69--79}.
\newblock
\urldef\tempurl%
\url{https://doi.org/10.18653/v1/N16-1009}
\showDOI{\tempurl}


\bibitem[\protect\citeauthoryear{Devlin, Chang, Lee, and Toutanova}{Devlin
  et~al\mbox{.}}{2018}]%
        {BERT}
\bibfield{author}{\bibinfo{person}{Jacob Devlin}, \bibinfo{person}{Ming-Wei
  Chang}, \bibinfo{person}{Kenton Lee}, {and} \bibinfo{person}{Kristina
  Toutanova}.} \bibinfo{year}{2018}\natexlab{}.
\newblock \bibinfo{title}{BERT: Pre-training of Deep Bidirectional Transformers
  for Language Understanding}.
\newblock
\newblock
\urldef\tempurl%
\url{http://arxiv.org/abs/1810.04805}
\showURL{%
\tempurl}
\newblock
\shownote{cite arxiv:1810.04805Comment: 13 pages.}


\bibitem[\protect\citeauthoryear{Dhoolia, Kumar, Contractor, and Joshi}{Dhoolia
  et~al\mbox{.}}{2021}]%
        {AAAI21}
\bibfield{author}{\bibinfo{person}{Pankaj Dhoolia}, \bibinfo{person}{Vineet
  Kumar}, \bibinfo{person}{Danish Contractor}, {and} \bibinfo{person}{Sachindra
  Joshi}.} \bibinfo{year}{2021}\natexlab{}.
\newblock \showarticletitle{Bootstrapping Dialog Models from Human to Human
  Conversation Logs}. In \bibinfo{booktitle}{\emph{Thirty-Fifth {AAAI}
  Conference on Artificial Intelligence, {AAAI} 2021, Thirty-Third Conference
  on Innovative Applications of Artificial Intelligence, {IAAI} 2021, The
  Eleventh Symposium on Educational Advances in Artificial Intelligence, {EAAI}
  2021, Virtual Event, February 2-9, 2021}}. \bibinfo{publisher}{{AAAI} Press},
  \bibinfo{pages}{16024--16025}.
\newblock
\urldef\tempurl%
\url{https://ojs.aaai.org/index.php/AAAI/article/view/18000}
\showURL{%
\tempurl}


\bibitem[\protect\citeauthoryear{Fadnis, Mills, Ganhotra, Roitman, Pandey,
  Cohen, Mass, Erera, Gunasekara, Contractor, Patel, Liao, Joshi, Lastras, and
  Konopnicki}{Fadnis et~al\mbox{.}}{2020}]%
        {AgentAssist}
\bibfield{author}{\bibinfo{person}{Kshitij~P. Fadnis},
  \bibinfo{person}{Nathaniel Mills}, \bibinfo{person}{Jatin Ganhotra},
  \bibinfo{person}{Haggai Roitman}, \bibinfo{person}{Gaurav Pandey},
  \bibinfo{person}{Doron Cohen}, \bibinfo{person}{Yosi Mass},
  \bibinfo{person}{Shai Erera}, \bibinfo{person}{R.~Chulaka Gunasekara},
  \bibinfo{person}{Danish Contractor}, \bibinfo{person}{Siva~Sankalp Patel},
  \bibinfo{person}{Q.~Vera Liao}, \bibinfo{person}{Sachindra Joshi},
  \bibinfo{person}{Luis~A. Lastras}, {and} \bibinfo{person}{David Konopnicki}.}
  \bibinfo{year}{2020}\natexlab{}.
\newblock \showarticletitle{Agent Assist through Conversation Analysis}. In
  \bibinfo{booktitle}{\emph{Proceedings of the 2020 Conference on Empirical
  Methods in Natural Language Processing: System Demonstrations, {EMNLP} 2020 -
  Demos, Online, November 16-20, 2020}}, \bibfield{editor}{\bibinfo{person}{Qun
  Liu} {and} \bibinfo{person}{David Schlangen}} (Eds.).
  \bibinfo{publisher}{Association for Computational Linguistics},
  \bibinfo{pages}{151--157}.
\newblock
\urldef\tempurl%
\url{https://doi.org/10.18653/v1/2020.emnlp-demos.20}
\showDOI{\tempurl}


\bibitem[\protect\citeauthoryear{Gu, Cho, Ha, and Kim}{Gu
  et~al\mbox{.}}{2018}]%
        {gu2018dialogwae}
\bibfield{author}{\bibinfo{person}{Xiaodong Gu}, \bibinfo{person}{Kyunghyun
  Cho}, \bibinfo{person}{Jung-Woo Ha}, {and} \bibinfo{person}{Sunghun Kim}.}
  \bibinfo{year}{2018}\natexlab{}.
\newblock \showarticletitle{DialogWAE: Multimodal Response Generation with
  Conditional Wasserstein Auto-Encoder}. In
  \bibinfo{booktitle}{\emph{International Conference on Learning
  Representations}}.
\newblock


\bibitem[\protect\citeauthoryear{Hershey and Olsen}{Hershey and Olsen}{2007}]%
        {kl_gmm_approx}
\bibfield{author}{\bibinfo{person}{John~R Hershey} {and}
  \bibinfo{person}{Peder~A Olsen}.} \bibinfo{year}{2007}\natexlab{}.
\newblock \showarticletitle{Approximating the Kullback Leibler divergence
  between Gaussian mixture models}. In \bibinfo{booktitle}{\emph{2007 IEEE
  International Conference on Acoustics, Speech and Signal
  Processing-ICASSP'07}}, Vol.~\bibinfo{volume}{4}. IEEE,
  \bibinfo{pages}{IV--317}.
\newblock


\bibitem[\protect\citeauthoryear{Humeau, Shuster, Lachaux, and Weston}{Humeau
  et~al\mbox{.}}{2019}]%
        {PolyEncoder}
\bibfield{author}{\bibinfo{person}{Samuel Humeau}, \bibinfo{person}{Kurt
  Shuster}, \bibinfo{person}{Marie-Anne Lachaux}, {and} \bibinfo{person}{Jason
  Weston}.} \bibinfo{year}{2019}\natexlab{}.
\newblock \showarticletitle{Poly-encoders: Transformer Architectures and
  Pre-training Strategies for Fast and Accurate Multi-sentence Scoring}.
\newblock \bibinfo{journal}{\emph{arXiv: Computation and Language}}
  (\bibinfo{year}{2019}).
\newblock


\bibitem[\protect\citeauthoryear{Ji, Lu, and Li}{Ji et~al\mbox{.}}{2014}]%
        {Ji2014AnIR}
\bibfield{author}{\bibinfo{person}{Zongcheng Ji}, \bibinfo{person}{Zhengdong
  Lu}, {and} \bibinfo{person}{Hang Li}.} \bibinfo{year}{2014}\natexlab{}.
\newblock \showarticletitle{An Information Retrieval Approach to Short Text
  Conversation}.
\newblock \bibinfo{journal}{\emph{CoRR}}  \bibinfo{volume}{abs/1408.6988}
  (\bibinfo{year}{2014}).
\newblock


\bibitem[\protect\citeauthoryear{Johnson, Douze, and J{\'e}gou}{Johnson
  et~al\mbox{.}}{2019}]%
        {faiss}
\bibfield{author}{\bibinfo{person}{Jeff Johnson}, \bibinfo{person}{Matthijs
  Douze}, {and} \bibinfo{person}{Herv{\'e} J{\'e}gou}.}
  \bibinfo{year}{2019}\natexlab{}.
\newblock \showarticletitle{Billion-scale similarity search with gpus}.
\newblock \bibinfo{journal}{\emph{IEEE Transactions on Big Data}}
  (\bibinfo{year}{2019}).
\newblock


\bibitem[\protect\citeauthoryear{Karpukhin, Oguz, Min, Lewis, Wu, Edunov, Chen,
  and Yih}{Karpukhin et~al\mbox{.}}{2020}]%
        {DPR}
\bibfield{author}{\bibinfo{person}{Vladimir Karpukhin}, \bibinfo{person}{Barlas
  Oguz}, \bibinfo{person}{Sewon Min}, \bibinfo{person}{Patrick Lewis},
  \bibinfo{person}{Ledell Wu}, \bibinfo{person}{Sergey Edunov},
  \bibinfo{person}{Danqi Chen}, {and} \bibinfo{person}{Wen-tau Yih}.}
  \bibinfo{year}{2020}\natexlab{}.
\newblock \showarticletitle{Dense Passage Retrieval for Open-Domain Question
  Answering}. In \bibinfo{booktitle}{\emph{Proceedings of the 2020 Conference
  on Empirical Methods in Natural Language Processing (EMNLP)}}.
  \bibinfo{publisher}{Association for Computational Linguistics},
  \bibinfo{address}{Online}, \bibinfo{pages}{6769--6781}.
\newblock
\urldef\tempurl%
\url{https://doi.org/10.18653/v1/2020.emnlp-main.550}
\showDOI{\tempurl}


\bibitem[\protect\citeauthoryear{Kenton and Toutanova}{Kenton and
  Toutanova}{2019}]%
        {kenton2019bert}
\bibfield{author}{\bibinfo{person}{Jacob Devlin Ming-Wei~Chang Kenton} {and}
  \bibinfo{person}{Lee~Kristina Toutanova}.} \bibinfo{year}{2019}\natexlab{}.
\newblock \showarticletitle{BERT: Pre-training of Deep Bidirectional
  Transformers for Language Understanding}. In
  \bibinfo{booktitle}{\emph{Proceedings of NAACL-HLT}}.
  \bibinfo{pages}{4171--4186}.
\newblock


\bibitem[\protect\citeauthoryear{Khattab and Zaharia}{Khattab and
  Zaharia}{2020}]%
        {khattab2020colbert}
\bibfield{author}{\bibinfo{person}{Omar Khattab} {and} \bibinfo{person}{Matei
  Zaharia}.} \bibinfo{year}{2020}\natexlab{}.
\newblock \showarticletitle{Colbert: Efficient and effective passage search via
  contextualized late interaction over bert}. In
  \bibinfo{booktitle}{\emph{Proceedings of the 43rd International ACM SIGIR
  conference on research and development in Information Retrieval}}.
  \bibinfo{pages}{39--48}.
\newblock


\bibitem[\protect\citeauthoryear{Lewis, Perez, Piktus, Petroni, Karpukhin,
  Goyal, Kuttler, Lewis, tau Yih, Rockt{\"a}schel, Riedel, and Kiela}{Lewis
  et~al\mbox{.}}{2020}]%
        {RAG}
\bibfield{author}{\bibinfo{person}{Patrick Lewis}, \bibinfo{person}{Ethan
  Perez}, \bibinfo{person}{Aleksandara Piktus}, \bibinfo{person}{Fabio
  Petroni}, \bibinfo{person}{Vladimir Karpukhin}, \bibinfo{person}{Naman
  Goyal}, \bibinfo{person}{Heinrich Kuttler}, \bibinfo{person}{Mike Lewis},
  \bibinfo{person}{Wen tau Yih}, \bibinfo{person}{Tim Rockt{\"a}schel},
  \bibinfo{person}{Sebastian Riedel}, {and} \bibinfo{person}{Douwe Kiela}.}
  \bibinfo{year}{2020}\natexlab{}.
\newblock \showarticletitle{Retrieval-Augmented Generation for
  Knowledge-Intensive NLP Tasks}.
\newblock \bibinfo{journal}{\emph{ArXiv}}  \bibinfo{volume}{abs/2005.11401}
  (\bibinfo{year}{2020}).
\newblock


\bibitem[\protect\citeauthoryear{Li, Galley, Brockett, Gao, and Dolan}{Li
  et~al\mbox{.}}{2016}]%
        {diversitypaper}
\bibfield{author}{\bibinfo{person}{Jiwei Li}, \bibinfo{person}{Michel Galley},
  \bibinfo{person}{Chris Brockett}, \bibinfo{person}{Jianfeng Gao}, {and}
  \bibinfo{person}{Bill Dolan}.} \bibinfo{year}{2016}\natexlab{}.
\newblock \showarticletitle{A Diversity-Promoting Objective Function for Neural
  Conversation Models}. In \bibinfo{booktitle}{\emph{{NAACL} {HLT} 2016, The
  2016 Conference of the North American Chapter of the Association for
  Computational Linguistics: Human Language Technologies, San Diego California,
  USA, June 12-17, 2016}}. \bibinfo{pages}{110--119}.
\newblock
\urldef\tempurl%
\url{http://aclweb.org/anthology/N/N16/N16-1014.pdf}
\showURL{%
\tempurl}


\bibitem[\protect\citeauthoryear{Liu, Wang, Wang, Ye, and Zhang}{Liu
  et~al\mbox{.}}{2021}]%
        {liu-etal-2021-quadrupletbert}
\bibfield{author}{\bibinfo{person}{Peiyang Liu}, \bibinfo{person}{Sen Wang},
  \bibinfo{person}{Xi Wang}, \bibinfo{person}{Wei Ye}, {and}
  \bibinfo{person}{Shikun Zhang}.} \bibinfo{year}{2021}\natexlab{}.
\newblock \showarticletitle{{Q}uadruplet{BERT}: An Efficient Model For
  Embedding-Based Large-Scale Retrieval}. In
  \bibinfo{booktitle}{\emph{Proceedings of the 2021 Conference of the North
  American Chapter of the Association for Computational Linguistics: Human
  Language Technologies}}. \bibinfo{publisher}{Association for Computational
  Linguistics}, \bibinfo{address}{Online}, \bibinfo{pages}{3734--3739}.
\newblock
\urldef\tempurl%
\url{https://doi.org/10.18653/v1/2021.naacl-main.292}
\showDOI{\tempurl}


\bibitem[\protect\citeauthoryear{Lowe, Pow, Serban, and Pineau}{Lowe
  et~al\mbox{.}}{2015}]%
        {lowe2015ubuntu}
\bibfield{author}{\bibinfo{person}{Ryan Lowe}, \bibinfo{person}{Nissan Pow},
  \bibinfo{person}{Iulian Serban}, {and} \bibinfo{person}{Joelle Pineau}.}
  \bibinfo{year}{2015}\natexlab{}.
\newblock \showarticletitle{The ubuntu dialogue corpus: A large dataset for
  research in unstructured multi-turn dialogue systems}.
\newblock \bibinfo{journal}{\emph{arXiv preprint arXiv:1506.08909}}
  (\bibinfo{year}{2015}).
\newblock


\bibitem[\protect\citeauthoryear{Lu, Jiao, and Zhang}{Lu et~al\mbox{.}}{2020}]%
        {lu2020twinbert}
\bibfield{author}{\bibinfo{person}{Wenhao Lu}, \bibinfo{person}{Jian Jiao},
  {and} \bibinfo{person}{Ruofei Zhang}.} \bibinfo{year}{2020}\natexlab{}.
\newblock \showarticletitle{Twinbert: Distilling knowledge to twin-structured
  bert models for efficient retrieval}.
\newblock \bibinfo{journal}{\emph{arXiv preprint arXiv:2002.06275}}
  (\bibinfo{year}{2020}).
\newblock


\bibitem[\protect\citeauthoryear{Luan, Eisenstein, Toutanova, and Collins}{Luan
  et~al\mbox{.}}{2021a}]%
        {RetrievalTACL}
\bibfield{author}{\bibinfo{person}{Yi Luan}, \bibinfo{person}{Jacob
  Eisenstein}, \bibinfo{person}{Kristina Toutanova}, {and}
  \bibinfo{person}{Michael Collins}.} \bibinfo{year}{2021}\natexlab{a}.
\newblock \showarticletitle{Sparse, Dense, and Attentional Representations for
  Text Retrieval}.
\newblock \bibinfo{journal}{\emph{Transactions of the Association for
  Computational Linguistics}}  \bibinfo{volume}{9} (\bibinfo{year}{2021}),
  \bibinfo{pages}{329--345}.
\newblock
\urldef\tempurl%
\url{https://doi.org/10.1162/tacl_a_00369}
\showDOI{\tempurl}


\bibitem[\protect\citeauthoryear{Luan, Eisenstein, Toutanova, and Collins}{Luan
  et~al\mbox{.}}{2021b}]%
        {MEBERT}
\bibfield{author}{\bibinfo{person}{Yi Luan}, \bibinfo{person}{Jacob
  Eisenstein}, \bibinfo{person}{Kristina Toutanova}, {and}
  \bibinfo{person}{Michael Collins}.} \bibinfo{year}{2021}\natexlab{b}.
\newblock \showarticletitle{Sparse, Dense, and Attentional Representations for
  Text Retrieval}.
\newblock \bibinfo{journal}{\emph{Transactions of the Association for
  Computational Linguistics}}  \bibinfo{volume}{9} (\bibinfo{year}{2021}),
  \bibinfo{pages}{329--345}.
\newblock
\urldef\tempurl%
\url{https://doi.org/10.1162/tacl_a_00369}
\showDOI{\tempurl}


\bibitem[\protect\citeauthoryear{Mishra, Madan, Pandey, and Contractor}{Mishra
  et~al\mbox{.}}{2021}]%
        {VRAG}
\bibfield{author}{\bibinfo{person}{Mayank Mishra}, \bibinfo{person}{Dhiraj
  Madan}, \bibinfo{person}{Gaurav Pandey}, {and} \bibinfo{person}{Danish
  Contractor}.} \bibinfo{year}{2021}\natexlab{}.
\newblock \showarticletitle{Variational Learning for Unsupervised Knowledge
  Grounded Dialogs}.
\newblock \bibinfo{journal}{\emph{CoRR}}  \bibinfo{volume}{abs/2112.00653}
  (\bibinfo{year}{2021}).
\newblock
\showeprint[arXiv]{2112.00653}
\urldef\tempurl%
\url{https://arxiv.org/abs/2112.00653}
\showURL{%
\tempurl}


\bibitem[\protect\citeauthoryear{Mohapatra, Pandey, Contractor, and
  Joshi}{Mohapatra et~al\mbox{.}}{2021}]%
        {MWozBiswesh}
\bibfield{author}{\bibinfo{person}{Biswesh Mohapatra}, \bibinfo{person}{Gaurav
  Pandey}, \bibinfo{person}{Danish Contractor}, {and}
  \bibinfo{person}{Sachindra Joshi}.} \bibinfo{year}{2021}\natexlab{}.
\newblock \showarticletitle{Simulated Chats for Building Dialog Systems:
  Learning to Generate Conversations from Instructions}. In
  \bibinfo{booktitle}{\emph{Findings of the Association for Computational
  Linguistics: {EMNLP} 2021, Virtual Event / Punta Cana, Dominican Republic,
  16-20 November, 2021}}, \bibfield{editor}{\bibinfo{person}{Marie{-}Francine
  Moens}, \bibinfo{person}{Xuanjing Huang}, \bibinfo{person}{Lucia Specia},
  {and} \bibinfo{person}{Scott~Wen{-}tau Yih}} (Eds.).
  \bibinfo{publisher}{Association for Computational Linguistics},
  \bibinfo{pages}{1190--1203}.
\newblock
\urldef\tempurl%
\url{https://doi.org/10.18653/v1/2021.findings-emnlp.103}
\showDOI{\tempurl}


\bibitem[\protect\citeauthoryear{Nogueira and Cho}{Nogueira and Cho}{2019}]%
        {BERTIR}
\bibfield{author}{\bibinfo{person}{Rodrigo Nogueira} {and}
  \bibinfo{person}{Kyunghyun Cho}.} \bibinfo{year}{2019}\natexlab{}.
\newblock \showarticletitle{Passage Re-ranking with {BERT}}.
\newblock \bibinfo{journal}{\emph{CoRR}}  \bibinfo{volume}{abs/1901.04085}
  (\bibinfo{year}{2019}).
\newblock
\showeprint[arXiv]{1901.04085}
\urldef\tempurl%
\url{http://arxiv.org/abs/1901.04085}
\showURL{%
\tempurl}


\bibitem[\protect\citeauthoryear{Papineni, Roukos, Ward, and Zhu}{Papineni
  et~al\mbox{.}}{2002}]%
        {papineni2002bleu}
\bibfield{author}{\bibinfo{person}{Kishore Papineni}, \bibinfo{person}{Salim
  Roukos}, \bibinfo{person}{Todd Ward}, {and} \bibinfo{person}{Wei-Jing Zhu}.}
  \bibinfo{year}{2002}\natexlab{}.
\newblock \showarticletitle{BLEU: a method for automatic evaluation of machine
  translation}. In \bibinfo{booktitle}{\emph{Proceedings of the 40th annual
  meeting on association for computational linguistics}}. Association for
  Computational Linguistics, \bibinfo{pages}{311--318}.
\newblock


\bibitem[\protect\citeauthoryear{Park, Cho, and Kim}{Park
  et~al\mbox{.}}{2018}]%
        {park2018hierarchical}
\bibfield{author}{\bibinfo{person}{Yookoon Park}, \bibinfo{person}{Jaemin Cho},
  {and} \bibinfo{person}{Gunhee Kim}.} \bibinfo{year}{2018}\natexlab{}.
\newblock \showarticletitle{A Hierarchical Latent Structure for Variational
  Conversation Modeling}. In \bibinfo{booktitle}{\emph{Proceedings of the 2018
  Conference of the North American Chapter of the Association for Computational
  Linguistics: Human Language Technologies, Volume 1 (Long Papers)}}.
  \bibinfo{pages}{1792--1801}.
\newblock


\bibitem[\protect\citeauthoryear{Qian, Feng, Wen, and Chua}{Qian
  et~al\mbox{.}}{2021}]%
        {GaussianEMB2}
\bibfield{author}{\bibinfo{person}{Chen Qian}, \bibinfo{person}{Fuli Feng},
  \bibinfo{person}{Lijie Wen}, {and} \bibinfo{person}{Tat-Seng Chua}.}
  \bibinfo{year}{2021}\natexlab{}.
\newblock \showarticletitle{Conceptualized and Contextualized Gaussian
  Embedding}.
\newblock \bibinfo{journal}{\emph{Proceedings of the AAAI Conference on
  Artificial Intelligence}} \bibinfo{volume}{35}, \bibinfo{number}{15}
  (\bibinfo{date}{May} \bibinfo{year}{2021}), \bibinfo{pages}{13683--13691}.
\newblock
\urldef\tempurl%
\url{https://ojs.aaai.org/index.php/AAAI/article/view/17613}
\showURL{%
\tempurl}


\bibitem[\protect\citeauthoryear{Qu, Yang, Chen, Qiu, Croft, and Iyyer}{Qu
  et~al\mbox{.}}{2020}]%
        {OrQUAC}
\bibfield{author}{\bibinfo{person}{Chen Qu}, \bibinfo{person}{Liu Yang},
  \bibinfo{person}{Cen Chen}, \bibinfo{person}{Minghui Qiu},
  \bibinfo{person}{W.~Bruce Croft}, {and} \bibinfo{person}{Mohit Iyyer}.}
  \bibinfo{year}{2020}\natexlab{}.
\newblock \bibinfo{booktitle}{\emph{Open-Retrieval Conversational Question
  Answering}}.
\newblock \bibinfo{publisher}{Association for Computing Machinery},
  \bibinfo{address}{New York, NY, USA}, \bibinfo{pages}{539–548}.
\newblock
\showISBNx{9781450380164}
\urldef\tempurl%
\url{https://doi.org/10.1145/3397271.3401110}
\showURL{%
\tempurl}


\bibitem[\protect\citeauthoryear{Raghu, Gupta, and Mausam}{Raghu
  et~al\mbox{.}}{2021}]%
        {DineshTACL}
\bibfield{author}{\bibinfo{person}{Dinesh Raghu}, \bibinfo{person}{Nikhil
  Gupta}, {and} \bibinfo{person}{Mausam}.} \bibinfo{year}{2021}\natexlab{}.
\newblock \showarticletitle{Unsupervised Learning of {KB} Queries in
  Task-Oriented Dialogs}.
\newblock \bibinfo{journal}{\emph{Trans. Assoc. Comput. Linguistics}}
  \bibinfo{volume}{9} (\bibinfo{year}{2021}), \bibinfo{pages}{374--390}.
\newblock
\urldef\tempurl%
\url{https://transacl.org/ojs/index.php/tacl/article/view/2515}
\showURL{%
\tempurl}


\bibitem[\protect\citeauthoryear{Reddy, Contractor, Raghu, and Joshi}{Reddy
  et~al\mbox{.}}{2019}]%
        {Multilevel}
\bibfield{author}{\bibinfo{person}{Revanth Reddy}, \bibinfo{person}{Danish
  Contractor}, \bibinfo{person}{Dinesh Raghu}, {and} \bibinfo{person}{Sachindra
  Joshi}.} \bibinfo{year}{2019}\natexlab{}.
\newblock \showarticletitle{Multi-Level Memory for Task Oriented Dialogs}. In
  \bibinfo{booktitle}{\emph{Proceedings of the 2019 Conference of the North
  American Chapter of the Association for Computational Linguistics: Human
  Language Technologies, {NAACL-HLT} 2019, Minneapolis, MN, USA, June 2-7,
  2019, Volume 1 (Long and Short Papers)}},
  \bibfield{editor}{\bibinfo{person}{Jill Burstein}, \bibinfo{person}{Christy
  Doran}, {and} \bibinfo{person}{Thamar Solorio}} (Eds.).
  \bibinfo{publisher}{Association for Computational Linguistics},
  \bibinfo{pages}{3744--3754}.
\newblock
\urldef\tempurl%
\url{https://doi.org/10.18653/v1/n19-1375}
\showDOI{\tempurl}


\bibitem[\protect\citeauthoryear{Reimers and Gurevych}{Reimers and
  Gurevych}{2019}]%
        {SBERT}
\bibfield{author}{\bibinfo{person}{Nils Reimers} {and} \bibinfo{person}{Iryna
  Gurevych}.} \bibinfo{year}{2019}\natexlab{}.
\newblock \showarticletitle{Sentence-{BERT}: Sentence Embeddings using
  {S}iamese {BERT}-Networks}. In \bibinfo{booktitle}{\emph{Proceedings of the
  2019 Conference on Empirical Methods in Natural Language Processing and the
  9th International Joint Conference on Natural Language Processing
  (EMNLP-IJCNLP)}}. \bibinfo{publisher}{Association for Computational
  Linguistics}, \bibinfo{address}{Hong Kong, China},
  \bibinfo{pages}{3982--3992}.
\newblock
\urldef\tempurl%
\url{https://doi.org/10.18653/v1/D19-1410}
\showDOI{\tempurl}


\bibitem[\protect\citeauthoryear{Robertson}{Robertson}{2009}]%
        {BM25}
\bibfield{author}{\bibinfo{person}{S. Robertson}.}
  \bibinfo{year}{2009}\natexlab{}.
\newblock \showarticletitle{{The Probabilistic Relevance Framework: BM25 and
  Beyond}}.
\newblock \bibinfo{journal}{\emph{Foundations and Trends{\textregistered} in
  Information Retrieval}} \bibinfo{volume}{3}, \bibinfo{number}{4}
  (\bibinfo{year}{2009}), \bibinfo{pages}{333--389}.
\newblock
\urldef\tempurl%
\url{http://scholar.google.de/scholar.bib?q=info:U4l9kCVIssAJ:scholar.google.com/&output=citation&hl=de&as_sdt=2000&as_vis=1&ct=citation&cd=1}
\showURL{%
\tempurl}


\bibitem[\protect\citeauthoryear{Serban, Sordoni, Bengio, Courville, and
  Pineau}{Serban et~al\mbox{.}}{2016}]%
        {Serban2016BuildingED}
\bibfield{author}{\bibinfo{person}{Iulian Serban}, \bibinfo{person}{Alessandro
  Sordoni}, \bibinfo{person}{Yoshua Bengio}, \bibinfo{person}{Aaron~C.
  Courville}, {and} \bibinfo{person}{Joelle Pineau}.}
  \bibinfo{year}{2016}\natexlab{}.
\newblock \showarticletitle{Building End-To-End Dialogue Systems Using
  Generative Hierarchical Neural Network Models}. In
  \bibinfo{booktitle}{\emph{AAAI}}.
\newblock


\bibitem[\protect\citeauthoryear{Serban, Sordoni, Lowe, Charlin, Pineau,
  Courville, and Bengio}{Serban et~al\mbox{.}}{2017}]%
        {serban2017hierarchical}
\bibfield{author}{\bibinfo{person}{Iulian~Vlad Serban},
  \bibinfo{person}{Alessandro Sordoni}, \bibinfo{person}{Ryan Lowe},
  \bibinfo{person}{Laurent Charlin}, \bibinfo{person}{Joelle Pineau},
  \bibinfo{person}{Aaron~C Courville}, {and} \bibinfo{person}{Yoshua Bengio}.}
  \bibinfo{year}{2017}\natexlab{}.
\newblock \showarticletitle{A Hierarchical Latent Variable Encoder-Decoder
  Model for Generating Dialogues.}. In \bibinfo{booktitle}{\emph{AAAI}}.
  \bibinfo{pages}{3295--3301}.
\newblock


\bibitem[\protect\citeauthoryear{Sohn}{Sohn}{2016}]%
        {sohn2016improved}
\bibfield{author}{\bibinfo{person}{Kihyuk Sohn}.}
  \bibinfo{year}{2016}\natexlab{}.
\newblock \showarticletitle{Improved deep metric learning with multi-class
  n-pair loss objective}. In \bibinfo{booktitle}{\emph{Advances in neural
  information processing systems}}. \bibinfo{pages}{1857--1865}.
\newblock


\bibitem[\protect\citeauthoryear{Sordoni, Galley, Auli, Brockett, Ji, Mitchell,
  Nie, Gao, and Dolan}{Sordoni et~al\mbox{.}}{2015}]%
        {Sordoni2015ANN}
\bibfield{author}{\bibinfo{person}{Alessandro Sordoni}, \bibinfo{person}{Michel
  Galley}, \bibinfo{person}{Michael Auli}, \bibinfo{person}{Chris Brockett},
  \bibinfo{person}{Yangfeng Ji}, \bibinfo{person}{Margaret Mitchell},
  \bibinfo{person}{Jian-Yun Nie}, \bibinfo{person}{Jianfeng Gao}, {and}
  \bibinfo{person}{William~B. Dolan}.} \bibinfo{year}{2015}\natexlab{}.
\newblock \showarticletitle{A Neural Network Approach to Context-Sensitive
  Generation of Conversational Responses}. In
  \bibinfo{booktitle}{\emph{HLT-NAACL}}.
\newblock


\bibitem[\protect\citeauthoryear{Sun, Zhao, Chen, Schroff, Adam, and Liu}{Sun
  et~al\mbox{.}}{2020}]%
        {poseprob}
\bibfield{author}{\bibinfo{person}{Jennifer~J. Sun}, \bibinfo{person}{Jiaping
  Zhao}, \bibinfo{person}{Liang{-}Chieh Chen}, \bibinfo{person}{Florian
  Schroff}, \bibinfo{person}{Hartwig Adam}, {and} \bibinfo{person}{Ting Liu}.}
  \bibinfo{year}{2020}\natexlab{}.
\newblock \showarticletitle{View-Invariant Probabilistic Embedding for Human
  Pose}. In \bibinfo{booktitle}{\emph{Computer Vision - {ECCV} 2020 - 16th
  European Conference, Glasgow, UK, August 23-28, 2020, Proceedings, Part {V}}}
  \emph{(\bibinfo{series}{Lecture Notes in Computer Science},
  Vol.~\bibinfo{volume}{12350})}, \bibfield{editor}{\bibinfo{person}{Andrea
  Vedaldi}, \bibinfo{person}{Horst Bischof}, \bibinfo{person}{Thomas Brox},
  {and} \bibinfo{person}{Jan{-}Michael Frahm}} (Eds.).
  \bibinfo{publisher}{Springer}, \bibinfo{pages}{53--70}.
\newblock
\urldef\tempurl%
\url{https://doi.org/10.1007/978-3-030-58558-7\_4}
\showDOI{\tempurl}


\bibitem[\protect\citeauthoryear{Tao, Feng, Yan, Wu, and Jiang}{Tao
  et~al\mbox{.}}{2021}]%
        {IJCAI21}
\bibfield{author}{\bibinfo{person}{Chongyang Tao}, \bibinfo{person}{Jiazhan
  Feng}, \bibinfo{person}{Rui Yan}, \bibinfo{person}{Wei Wu}, {and}
  \bibinfo{person}{Daxin Jiang}.} \bibinfo{year}{2021}\natexlab{}.
\newblock \showarticletitle{A survey on response selection for retrieval-based
  dialogues}. In \bibinfo{booktitle}{\emph{IJCAI}}.
\newblock


\bibitem[\protect\citeauthoryear{Vakili~Tahami, Ghajar, and
  Shakery}{Vakili~Tahami et~al\mbox{.}}{2020}]%
        {distillconvIR}
\bibfield{author}{\bibinfo{person}{Amir Vakili~Tahami}, \bibinfo{person}{Kamyar
  Ghajar}, {and} \bibinfo{person}{Azadeh Shakery}.}
  \bibinfo{year}{2020}\natexlab{}.
\newblock \bibinfo{booktitle}{\emph{Distilling Knowledge for Fast
  Retrieval-Based Chat-Bots}}.
\newblock \bibinfo{publisher}{Association for Computing Machinery},
  \bibinfo{address}{New York, NY, USA}, \bibinfo{pages}{2081–2084}.
\newblock
\showISBNx{9781450380164}
\urldef\tempurl%
\url{https://doi.org/10.1145/3397271.3401296}
\showURL{%
\tempurl}


\bibitem[\protect\citeauthoryear{Vinyals and Le}{Vinyals and Le}{2015}]%
        {Vinyals2015ANC}
\bibfield{author}{\bibinfo{person}{Oriol Vinyals} {and}
  \bibinfo{person}{Quoc~V. Le}.} \bibinfo{year}{2015}\natexlab{}.
\newblock \showarticletitle{A Neural Conversational Model}.
\newblock \bibinfo{journal}{\emph{CoRR}}  \bibinfo{volume}{abs/1506.05869}
  (\bibinfo{year}{2015}).
\newblock


\bibitem[\protect\citeauthoryear{Wu, Wu, Xing, Xu, Li, and Zhou}{Wu
  et~al\mbox{.}}{2017a}]%
        {Wu2017ASM}
\bibfield{author}{\bibinfo{person}{Yu Wu}, \bibinfo{person}{Wei Wu},
  \bibinfo{person}{Chen Xing}, \bibinfo{person}{Can Xu},
  \bibinfo{person}{Zhoujun Li}, {and} \bibinfo{person}{Ming Zhou}.}
  \bibinfo{year}{2017}\natexlab{a}.
\newblock \showarticletitle{A Sequential Matching Framework for Multi-turn
  Response Selection in Retrieval-based Chatbots}.
\newblock \bibinfo{journal}{\emph{CoRR}}  \bibinfo{volume}{abs/1710.11344}
  (\bibinfo{year}{2017}).
\newblock


\bibitem[\protect\citeauthoryear{Wu, Wu, Zhou, and Li}{Wu
  et~al\mbox{.}}{2017b}]%
        {Wu2017SequentialMN}
\bibfield{author}{\bibinfo{person}{Yu Wu}, \bibinfo{person}{Wei Wu},
  \bibinfo{person}{Ming Zhou}, {and} \bibinfo{person}{Zhoujun Li}.}
  \bibinfo{year}{2017}\natexlab{b}.
\newblock \showarticletitle{Sequential Match Network: A New Architecture for
  Multi-turn Response Selection in Retrieval-based Chatbots}. In
  \bibinfo{booktitle}{\emph{ACL}}.
\newblock


\bibitem[\protect\citeauthoryear{Xiong, Xiong, Li, Tang, Liu, Bennett, Ahmed,
  and Overwijk}{Xiong et~al\mbox{.}}{2021}]%
        {ANNICLR}
\bibfield{author}{\bibinfo{person}{Lee Xiong}, \bibinfo{person}{Chenyan Xiong},
  \bibinfo{person}{Ye Li}, \bibinfo{person}{Kwok{-}Fung Tang},
  \bibinfo{person}{Jialin Liu}, \bibinfo{person}{Paul~N. Bennett},
  \bibinfo{person}{Junaid Ahmed}, {and} \bibinfo{person}{Arnold Overwijk}.}
  \bibinfo{year}{2021}\natexlab{}.
\newblock \showarticletitle{Approximate Nearest Neighbor Negative Contrastive
  Learning for Dense Text Retrieval}. In \bibinfo{booktitle}{\emph{9th
  International Conference on Learning Representations, {ICLR} 2021, Virtual
  Event, Austria, May 3-7, 2021}}. \bibinfo{publisher}{OpenReview.net}.
\newblock
\urldef\tempurl%
\url{https://openreview.net/forum?id=zeFrfgyZln}
\showURL{%
\tempurl}


\bibitem[\protect\citeauthoryear{Yan, Song, and Wu}{Yan et~al\mbox{.}}{2016}]%
        {Yan2016LearningTR}
\bibfield{author}{\bibinfo{person}{Rui Yan}, \bibinfo{person}{Yiping Song},
  {and} \bibinfo{person}{Hua Wu}.} \bibinfo{year}{2016}\natexlab{}.
\newblock \showarticletitle{Learning to Respond with Deep Neural Networks for
  Retrieval-Based Human-Computer Conversation System}. In
  \bibinfo{booktitle}{\emph{SIGIR}}.
\newblock


\bibitem[\protect\citeauthoryear{Yu, Liu, Xiong, Feng, and Liu}{Yu
  et~al\mbox{.}}{2021}]%
        {ConvDR}
\bibfield{author}{\bibinfo{person}{Shi Yu}, \bibinfo{person}{Zhenghao Liu},
  \bibinfo{person}{Chenyan Xiong}, \bibinfo{person}{Tao Feng}, {and}
  \bibinfo{person}{Zhiyuan Liu}.} \bibinfo{year}{2021}\natexlab{}.
\newblock \bibinfo{booktitle}{\emph{Few-Shot Conversational Dense Retrieval}}.
\newblock \bibinfo{publisher}{Association for Computing Machinery},
  \bibinfo{address}{New York, NY, USA}, \bibinfo{pages}{829–838}.
\newblock
\showISBNx{9781450380379}
\urldef\tempurl%
\url{https://doi.org/10.1145/3404835.3462856}
\showURL{%
\tempurl}


\bibitem[\protect\citeauthoryear{Zhang, Sun, Galley, Chen, Brockett, Gao, Gao,
  Liu, and Dolan}{Zhang et~al\mbox{.}}{2020}]%
        {zhang2020dialogpt}
\bibfield{author}{\bibinfo{person}{Yizhe Zhang}, \bibinfo{person}{Siqi Sun},
  \bibinfo{person}{Michel Galley}, \bibinfo{person}{Yen-Chun Chen},
  \bibinfo{person}{Chris Brockett}, \bibinfo{person}{Xiang Gao},
  \bibinfo{person}{Jianfeng Gao}, \bibinfo{person}{Jingjing Liu}, {and}
  \bibinfo{person}{William~B Dolan}.} \bibinfo{year}{2020}\natexlab{}.
\newblock \showarticletitle{DIALOGPT: Large-Scale Generative Pre-training for
  Conversational Response Generation}. In \bibinfo{booktitle}{\emph{Proceedings
  of the 58th Annual Meeting of the Association for Computational Linguistics:
  System Demonstrations}}. \bibinfo{pages}{270--278}.
\newblock


\end{thebibliography}

\end{document}